\documentclass[10pt,twocolumn,letterpaper]{article}

\usepackage{iccv}
\usepackage{times}
\usepackage{epsfig}
\usepackage{graphicx}
\usepackage{amsmath}
\usepackage{amssymb}

\newtheorem{theorem}{Theorem}[section]
\newtheorem{corollary}[theorem]{Corollary}
\newtheorem{lemma}[theorem]{Lemma}
\usepackage{color, colortbl}
\definecolor{Gray}{rgb}{0.9,0.9,1}
\usepackage{algorithm}
\usepackage{multirow}
 \usepackage{subfig}
 \usepackage{float}
 \usepackage{nccmath}
 \usepackage{hhline}
 \usepackage{booktabs}

\usepackage[pagebackref=true,breaklinks=true,letterpaper=true,colorlinks,bookmarks=false]{hyperref}

\iccvfinalcopy 

\ificcvfinal\fi

\begin{document}

\title{CoDeC: Communication-Efficient Decentralized Continual Learning}

\author{Sakshi Choudhary,
Sai Aparna Aketi, Gobinda Saha and Kaushik Roy\\
Purdue University, West Lafayette, IN 47907, USA\\
{\tt\small \{choudh23, saketi, gsaha, kaushik\}@purdue.edu}
}

\maketitle

\ificcvfinal\thispagestyle{empty}\fi
\begin{abstract}
Training at the edge utilizes continuously evolving data generated at different locations.~Privacy concerns \mbox{prohibit the} co-location of this spatially as well as temporally distributed data, deeming it crucial to design training algorithms that enable efficient continual learning over decentralized private data. 
Decentralized learning \mbox{allows} serverless training with spatially distributed data.~A fundamental barrier in such distributed learning is the high bandwidth cost of communicating model updates between agents.~Moreover, existing works under this training paradigm are not inherently suitable for learning a temporal sequence of tasks while retaining the previously acquired knowledge.
In this work, we propose  CoDeC, a novel communication-efficient decentralized continual learning algorithm which addresses these challenges.~We mitigate catastrophic forgetting while learning a task sequence in a decentralized learning setup by combining orthogonal \mbox{gradient} projection with gossip averaging across decentralized agents.
Further, CoDeC includes a novel lossless communication compression scheme based on the gradient subspaces. We express layer-wise gradients as a linear combination of the basis vectors of these gradient subspaces and communicate the associated coefficients. 
We theoretically analyze the convergence rate for our algorithm and demonstrate through an extensive set of experiments that CoDeC successfully learns distributed continual tasks with minimal forgetting.~The proposed compression scheme results in up to 4.8$\times$ reduction in communication costs with iso-performance as the full communication baseline.
\end{abstract}

\section{Introduction}

Deep neural networks have demonstrated exceptional performance for many visual recognition tasks over the past decade. This has been fueled by the explosive growth of available training data and powerful computing resources. Edge devices such as smartphones, drones, and Internet-of-Things (IoT) sensors contribute towards generating this massive amount of data \cite{shi2020communication}. Interestingly, this data is spatially distributed, while continuously evolving over time. 
Large-scale deep neural network training has traditionally relied upon the availability of humongous amount of data at a central server. This mainly poses three challenges: (1) high network bandwidth requirements to collect this dispersed data from numerous learning agents, (2) data privacy concerns for locally-generated data accessed by the central server and (3) adapting to changing data distributions without expensive training from the scratch.
This motivates the need for learning algorithms to enable efficient distributed training by utilizing spatially and temporally distributed (i.e. non-stationary) data. 

\begin{figure*}
\centering
  \includegraphics[width=1\textwidth]{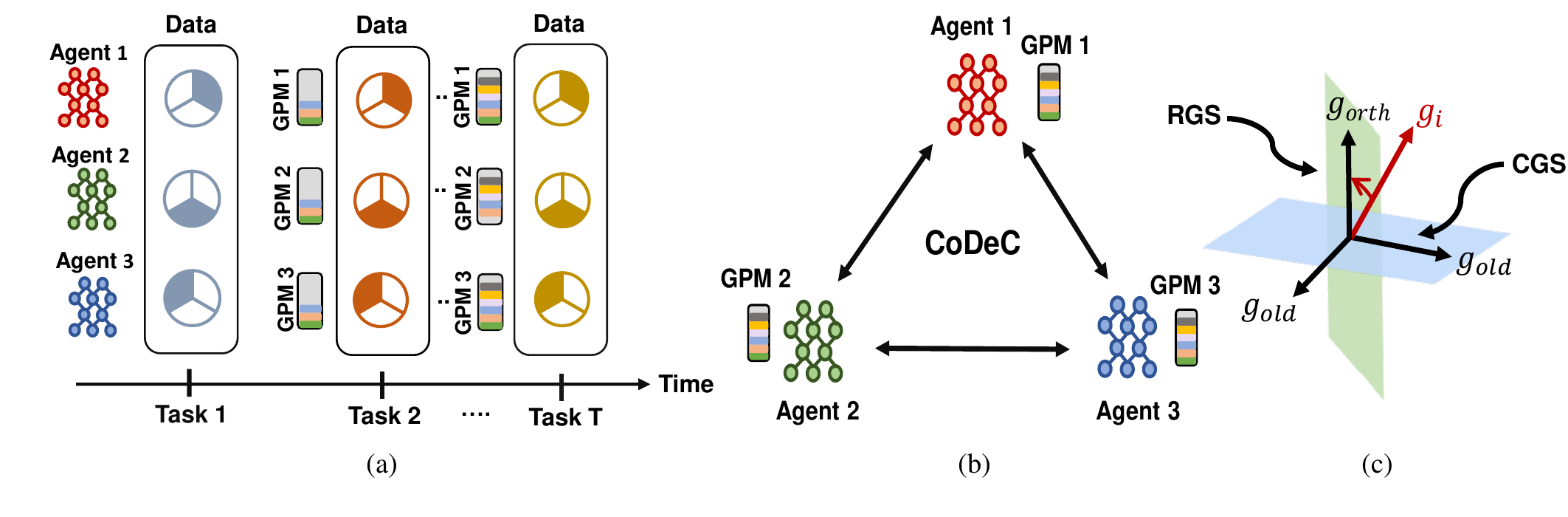}
  \vspace{-10mm}
  \caption{An overview of CoDeC. (a) Data for each incoming task is independently and identically distributed (IID) over the decentralized agents. Each agent has a GPM (Gradient Projection Memory) which is updated after learning each task. (b) Based on the sparse graph topology, the agents communicate coefficients associated with the model updates at each training iteration. (c) GPM partitions each layer's subspace into two orthogonal subspaces.}
  \label{fig:intro}
\end{figure*}

Centralized distributed learning (also known as federated learning) has emerged to train models over spatially distributed data without compromising on user privacy \cite{FedOpt}.
This approach relies upon a central parameter server to collect local model updates, process, and send the global updates back to the agents without accessing their local data. However, the central server may lead to a single point of failure and network bandwidth issues~\cite{sgp, flcooperate}. To address these concerns, several decentralized distributed learning algorithms have been developed ~\cite{bianchi,lan, DPSGD, sgp, momendpsgd}. Decentralized learning is a peer-to-peer learning paradigm, where agents communicate only with their neighbors without the need for a central parameter server. The aim of each learning agent is to learn a global generalized model by aggregating locally computed model updates shared by neighbors. The authors in \cite{DPSGD} propose Decentralized Parallel Stochastic Gradient Descent (DPSGD) by combining Stochastic Gradient Descent (SGD) with gossip averaging algorithm \cite{1272421} to show that decentralized algorithms perform similar to centralized algorithms on image classification datasets. 
However, decentralized learning algorithms are not inherently equipped to thrive in dynamic learning environments with a temporal sequence of changing data distributions. 

Continual learning algorithms act as a main tool to train models in dynamic learning environments.~Traditional DNN training utilizes gradient-based optimization methods like SGD and DPSGD \cite{DPSGD} which inherently update model parameters by minimizing the loss function with respect to the current data distribution. This results in overwriting of parameters learned for the previous task(s), leading to the phenomenon of catastrophic forgetting ~\cite{cat1,cat2}. Hence, continual learning techniques focus on learning consecutive tasks without forgetting the past acquired knowledge. Common approaches include dedicating sub-networks to tasks without any constraints on the network size \cite{pnn,den,l2g}, storing a subset of old data for rehearsal \cite{gem,icarl,agem}, freezing~\cite{packnet,space} or penalizing~\cite{ewc,oewc,mas} changes to parameters, or constraining~\cite{ogd,gpm} the important gradient directions.

In this paper, we propose CoDeC to enable serverless training with data distributed across space as well as time. To the best of our knowledge, this is the first work that demonstrates such a decentralized continual learning setup. Our algorithm has three components: (1) SGD combined with gossip averaging \cite{1272421} as shown in \cite{DPSGD} to learn with spatially distributed private data, (2) Gradient Projection Memory (GPM)~\cite{gpm} to continually learn a temporal task sequence with minimal forgetting and (3) a novel lossless communication compression scheme to reduce the bandwidth requirements of training models in this setup. We illustrate our setup in figures \ref{fig:intro}(a) and \ref{fig:intro}(b).

GPM \cite{gpm} partitions each layer's gradient space into two orthogonal subspaces: Core Gradient Space (CGS) and Residual Gradient Space (RGS) as shown in \ref{fig:intro}(c). Important gradient directions (CGS) for previous tasks are stored in gradient projection memory (GPM), and gradient updates for the new tasks are taken along RGS to minimize interference.  We find the basis vectors which span RGS and represent model updates as a linear combination of these vectors. We communicate the coefficients associated with these basis vectors instead of the model updates and achieve lossless communication compression. 
Further, theoretical insights into the achievable convergence rate for our algorithm prove that it is possible to achieve similar rates as the state-of-the-art decentralized learning approaches such as DPSGD \cite{DPSGD}. We provide empirical evidence of competitive performance by performing experiments over a variety of image-classification datasets and networks, as well as graph sizes and topologies. 

\noindent \textbf{Contributions:} The contributions of this work can be summarized as follows:
\begin{itemize}
\setlength\itemsep{-1mm}
\item We propose CoDeC, a communication-efficient decentralized continual learning algorithm which addresses a challenging problem: leveraging spatially and temporally distributed data to optimize a global model while preserving data privacy.
\item We introduce a novel lossless communication compression scheme based on gradient subspaces.
\item We theoretically show that our algorithm convergences at the rate of $O(1/\sqrt{NK})$, where $N$ is the number of agents and $K$ is the number of training iterations. This convergence rate is similar to the well-known decentralized learning methods \cite{DPSGD}.
\item Experiments over a variety of image-classification datasets, networks, graph sizes, and topologies demonstrate minimal forgetting and up to 4.8$\times$ reduction in communication costs with iso-performance relative to the full communication baseline.
\end{itemize}

\section{Related Work}
\subsection{Decentralized Learning}
Several works exist in the decentralized learning paradigm which enable distributing training without utilizing a central server~\cite{bianchi,lan, DPSGD, sgp, momendpsgd}. DPSGD~\cite{DPSGD} provides theoretical analysis for the convergence rate of decentralized learning algorithms, proving it to be similar to their centralized counterpart \cite{lsdnn}. The authors in \cite{sgp} extend DPSGD to directed and time-varying graphs. The authors in \cite{momendpsgd} propose Decentralized Momentum Stochastic Gradient Descent (DMSGD), which introduces momentum to DPSGD. In CoDeC, we utilize DPSGD~\cite{DPSGD} and modify it to send model updates instead of model parameters. Note, these existing works are not inherently equipped to learn a temporal task sequence without forgetting the past acquired knowledge. 

To reduce the communication overhead for decentralized learning, several error-compensation based communication compression techniques \cite{chocoSGD,ds,sp} have been explored. DeepSqueeze \cite{ds} is the first work that introduced error-compensated communication compression to decentralized training. The authors in \cite{sp} combined DeepSqueeze with Stochastic Gradient Push (SGP) \cite{sgp} to develop communication-efficient decentralized learning over directed and time-varying graphs. Choco-SGD\cite{chocoSGD} communicates compressed model updates rather than the model parameters and achieves better performance than \cite{ds}.
However, it is orthogonal to the compression scheme we present in this work and can be used in synergy with our approach. Moreover, all of the above-mentioned compression techniques are lossy and require additional hyperparameter tuning, unlike our proposed lossless compression scheme.
\subsection{Continual Learning}
The majority of continual learning works fall into three categories: network expansion, replay and regularization-based methods. Network expansion based methods overcome catastrophic forgetting by dedicating different model parameters to each task. With no constraints on the network size, \cite{pnn} adds new sub-networks for each new task while \cite{den} performs partial retraining and increases network capacity to account for newly acquired knowledge when necessary. Replay-based methods store training samples from the past tasks in the memory or synthesize old data from generative models for rehearsal \cite{agem,gem,icarl,genreplay}. GEM \cite{gem} and A-GEM \cite{agem} aim to minimize the loss on the current dataset as well as the episodic memory. When access to the past data is limited, \cite{genreplay} generates fake data to mimic training examples for rehearsal. Regularization-based methods tend to penalize changes to parameters ~\cite{ewc,oewc,mas}, or constrain gradient directions ~\cite{ogd,gpm}  important for previous tasks. All of these methods rely on the availability of the temporally distributed training data at a central location, and hence fail to be directly applicable to a distributed learning scenario. Network expansion based methods in a decentralized continual learning setup may give rise to model heterogeneity across agents over time, while replay-based methods can lead to privacy concerns. Thus, we explore regularization based methods like GPM \cite{gpm} and EWC \cite{ewc} in this work. We utilize GPM in CoDeC, and show superior performance than D-EWC, a decentralized continual learning baseline we implemented with EWC.

\subsection{Distributed Continual Learning}
FedWeIT \cite{yoon2021fed} tackled the problem of federated continual learning through decomposition of model parameters at each client into global and sparse local task-adaptive parameters. FLwF-2T \cite{usmandistil} developed a distillation-based method for class-incremental federated continual learning. Unlike our serverless training setup, these works utilize a central server to aggregate and send global updates to the agents. CoLLA \cite{colla} focused on multi-agent distributed lifelong learning and proposed a distributed optimization algorithm for a network of synchronous learning agents. 
However, it uses parametric models and is not directly applicable to modern deep neural networks.

\section{Methodology}
In this section, we formulate the problem and introduce our proposed decentralized continual learning setup with lossless compression scheme. 

\subsection{Problem Formulation}
In this work, we optimize a DNN model to learn from spatially and temporally distributed data. We consider a set of $N$ learning agents connected over a sparse communication topology. The communication topology is modeled as a graph $G=([N], \mathbf{W})$, where $\mathbf{W}$ is the mixing matrix indicating the graph's connectivity. In particular, $w_{ij}$ encodes the effect of agent $j$ on agent $i$, and $w_{ij}=0$ implies there is no communication link between the two agents. Note that there is no central server, and the agents can communicate only with their neighbors/peers. 

We consider a learning scenario where $T$ tasks are learned sequentially. Now, for any task $\tau\in \{1,..,T\}$, the corresponding dataset $\mathcal{D}_\tau$ is independently and identically distributed (IID) across the $N$ agents as $ \{\mathcal{D}_{\tau,1}, \mathcal{D}_{\tau,2}, \mathcal{D}_{\tau,3} ..... \mathcal{D}_{\tau,N} \} $.
For every task $\tau \in \{ 1,..,T \}$, we solve the optimization problem of minimizing global loss function $\mathcal{F}_\tau(\mathbf{x})$ distributed across the $N$ agents as given in equation~\ref{eq:1}. Here, $F_{\tau,i}(d_{\tau, i},\mathbf{x})$ is the local loss function per task at agent $i$ (e.g.\ cross-entropy loss)  and $f_{\tau,i}(\mathbf{x})$ is the expected value of $F_{\tau,i}(d_{\tau, i},\mathbf{x})$ over the dataset $\mathcal{D}_{\tau,i}$.
\vspace{-1.3mm}
\begin{equation}
\label{eq:1}
\begin{split}
    \min \limits_{\mathbf{x} \in \mathbb{R}^d} \mathcal{F}_\tau(\mathbf{x}) &= \frac{1}{N}\sum_{i=1}^N f_{\tau,i}(\mathbf{x}), \\
     where \hspace{3mm} f_{\tau,i}(\mathbf{x}) &= \mathbb{E}_{d_{\tau, i} \sim \mathcal{D}_{\tau, i}}[F_{\tau,i}(d_{\tau, i},\mathbf{x})] \hspace{2mm} \forall i
\end{split}
\vspace{-1.3mm}
\end{equation}

Decentralized optimization of this global loss function $\mathcal{F}_\tau(\mathbf{x})$ is based on the current dataset $\mathcal{D}_{\tau}$.  
A crucial challenge is to optimize $\mathcal{F}_\tau(\mathbf{x})$ such that the past information acquired from tasks $1,2,..,(\tau-1) $ is retained. Inspired by \cite{gpm}, we define a subspace that contains important gradient directions associated with all the past tasks and modify the local gradient updates of the current task to be orthogonal to this subspace i.e., to lie in RGS. This ensures minimal interference with the previously acquired knowledge, and hence minimal catastrophic forgetting. 

Typically, decentralized agents communicate the model parameters with their neighbors in each training iteration \cite{DPSGD}. Note that in the proposed algorithm the model updates lie in RGS, which is a smaller vector subspace compared to the entire gradient space. To utilize this property for enabling lossless communication compression (discussed in section \ref{lc}), we communicate model updates with neighbors similar to \cite{chocoSGD} rather than the model parameters.
\subsection{Approach}
We demonstrate the flow of CoDeC in Algorithm \ref{apx_alg:CoDeC}. 
All hyperparameters are synchronized between the agents at the beginning of the training.
\par Each agent $i$ computes the gradient update $\mathbf{g}^i= (\triangledown f_{\tau,i}(d_{\tau,i};\mathbf{x}^i))$ with respect to model parameters $\mathbf{x}^i$, evaluated on mini-batch $d_{\tau,i}$. We obtain $\tilde{\mathbf{g}}^i$, the orthogonal projection of the local gradients using GPM memory ${\mathcal{M}}$ (line 6, algorithm \ref{apx_alg:CoDeC}).
The parameters of each agent are updated using this $\tilde{\mathbf{g}}^i$ which ensures minimal forgetting. Then, each agent performs a gossip averaging step using $\mathbf{x}^i$ and $\hat{\mathbf{x}}^j$ (line 8, algorithm \ref{apx_alg:CoDeC}). $\hat{\mathbf{x}}^j$ represent the copies of $\mathbf{x}^j$ maintained by all the neighbors of agent $j$ and in general $\mathbf{x}^j=\hat{\mathbf{x}}^j$. 
The computed model updates (denoted by $\mathbf{q}_i^k$) lie in the RGS subspace spanned by the basis vectors contained in ${\mathbf{O}^l}$. Therefore, we express them as a linear combination of these basis vectors and find the associated coefficients, $\mathbf{c}^i$ to communicate with the neighbors as shown in line 10, algorithm \ref{apx_alg:CoDeC}. 
Upon receiving these coefficients, the agents reconstruct the neighbors' updates without any loss in information (line 13, algorithm \ref{apx_alg:CoDeC}). 
Communicating the coefficients ($\mathbf{c}^i$) leads to lossless compression, which we elaborate upon in section \ref{lc}. The local copy $\hat{\mathbf{x}}^j$ is updated using the reconstructed model updates $\mathbf{q}^j$ (line 14, algorithm \ref{apx_alg:CoDeC}). 
Note that our algorithm requires each agent to only store the sum of neighbors' models $\sum_{j \in \mathcal{N}(i)} w_{ij}\hat{\mathbf{x}}^j$ resulting in $O(1)$ memory overhead, independent of the number of neighbors.

At the end of each task, important gradient directions are obtained using a Singular Value Decomposition (SVD) representation of the input activations of each layer \cite{gpm}. These gradient directions are added as basis vectors to the CGS matrix $\mathcal{M}$ and subsequently removed from the RGS Matrix $\mathcal{O}$. Since we assume the data distribution for a given task across agents to be IID, we can compute SVD using input activations at any randomly chosen agent and communicate it to other agents iteratively using the communication graph.

\begin{algorithm}[t]
\textbf{Input:} Each agent $i \in [1,N]$ initializes model parameters $\mathbf{x}_0^{i}$, step size $\eta$, mixing matrix $\mathbf{W}=[w_{ij}]_{i,j \in [1,N]}$, $\hat{\mathbf{x}}_{(0)}^i\hspace{-1mm}=0$, $\mathbf{M}^l=[\hspace{1mm}]$ and $\mathbf{O}^l=[\mathbf{I}]$ for all layers $l= 1,2,...L$, GPM Memory $\mathcal{M}= \{(\mathbf{M}^l)^L_{l=1}\}$, RGS Matrix $\mathcal{O}=\{(\mathbf{O}^l)^L_{l=1}\}$, $\mathcal{N}(i)$:~neighbors of agent $i$ (including itself), $T$: total tasks, $K$: number of training iterations\\

Each agent simultaneously implements the 
T\text{\scriptsize RAIN}( ) procedure\\
\text{\small 1.}  \textbf{procedure} T\text{\scriptsize RAIN}( ) \\
\text{\small 2.}  \hspace{4mm}\textbf{for} $\tau=1,\hdots,T$ \textbf{do}\\
\text{\small 3.}  \hspace{8mm}\textbf{for} $k=0,1,\hdots,K-1$ \textbf{do}\\
\text{\small 4.}  \hspace*{12mm}$d_{\tau, i} \sim \mathcal{D}_{\tau,i}$\\
\text{\small 5.}  \hspace*{12mm}$\mathbf{g}^{i}_{k}=\nabla f_{\tau,i}(d_{\tau, i}; \mathbf{x}^i_{k}) $\\
\text{\small 6.}  \label{line6}\hspace*{12mm}$\tilde{\mathbf{g}}^i_{k}= \mathbf{g}^{i}_{k}-({\mathbf{M}^l}{(\mathbf{M}^l)^T})\mathbf{g}^{i}_{k}$ \space\space\# for each layer $l$\\
\text{\small 7.}  \hspace*{12mm}$\mathbf{x}_{(k+\frac{1}{2})}^i=\mathbf{x}_k^{i}- \eta \tilde{\mathbf{g}}^i_{k} $\\
\text{\small 8.}  \label{line8} \hspace*{12mm}$\mathbf{x}_{k+1}^{i}=\mathbf{x}_{(k+\frac{1}{2})}^i+ \sum_{j \in \mathcal{N}(i)} w_{ij}(\hat{\mathbf{x}}_k^{j}-\mathbf{x}_k^{i})$\\    
\text{\small 9.} \hspace*{12mm}$\mathbf{q}_{k}^i= \mathbf{x}^i_{k+1}-\mathbf{x}^i_{k}$\\
\text{\small 10.} \hspace*{11mm}$\mathbf{c}_{k}^i= {(\mathbf{O}^l)}^T\mathbf{q}_{k}^i$\\
\text{\small 11.} \hspace*{11mm} for each $j \in \mathcal{N}(i)$ \textbf{do}\\
\text{\small 12.}    \hspace*{15mm} Send $\mathbf{c}_{k}^i$ and receive $\mathbf{c}_{k}^j$\\
\text{\small 13.}     \hspace*{15mm} $\mathbf{q}_{k}^j={(\mathbf{O}^l)}\mathbf{c}_{k}^j$\\
\text{\small 14.}     \hspace*{15mm} $\hat{\mathbf{x}}^j_{(k+1)}= \mathbf{q}_{k}^j+\hat{\mathbf{x}}_k^{j}$\\
\text{\small 15.}    \hspace*{11mm}\textbf{end}\\   
\text{\small 16.} \hspace*{8mm}\textbf{end}\\
\hspace*{12mm} \# GPM Update  \\
\text{\small 17.}    \hspace*{8mm} $p= random(1,2,...N)$ \\
\text{\small 18.}    \hspace*{8mm} \textbf{if} $i==p$ \textbf{do} \\
\text{\small 19.}   \hspace*{12mm}Update $\mathbf{M}^l$, $\mathbf{O}^l$ for each layer $l \in L$\\
\text{\small 20.}   \hspace*{12mm}Update $\mathcal{M}=\{(\mathbf{M}^l)^L_{l=1}\}$\\
\text{\small 21.}    \hspace*{12mm}Update $\mathcal{O}=\{(\mathbf{O}^l)^L_{l=1}\}$\\
\text{\small 22.}     \hspace*{12mm}Send $\mathcal{M}$, $\mathcal{O}$ to all agents\\
\text{\small 23.}    \hspace*{8mm} \textbf{end}\\
\text{\small 24.}   \hspace*{4mm}\textbf{end}\\
\text{\small 25.} \textbf{return}
\caption{Communication-Efficient Decentralized Continual Learning (\textit{CoDeC})}
\label{apx_alg:CoDeC}
\end{algorithm}
\setlength{\textfloatsep}{10pt}

\subsection{Lossless Compression}
\label{lc}
\vspace{-0.5em}
Stochastic Gradient Descent (SGD) updates lie in the span of input data points \cite{zhang}. Leveraging this fact, GPM \cite{gpm} performs SVD on a representation matrix $\mathbf{R}_\tau^l$ and finds basis vectors corresponding to the most important gradient directions for the previous tasks. $\mathbf{R}_\tau^l$ is constructed by performing a forward pass of $n_s$ samples from the training dataset for task $\tau$ through the network and concatenating the input activations for each layer $l$ as shown in equation \ref{repres}. Subsequently, the SVD of representation, $\mathbf{R}_\tau^l$ in equation \ref{repres} is used to obtain the matrix $\mathbf{U}_{\tau}^l$ containing a set of orthonormal basis vectors which span the entire gradient space. 
\begin{equation}
\label{repres}
\begin{split}
    \mathbf{R}_\tau^l= [x_{1,\tau}^l, x_{2,\tau}^l.., x_{n_s,\tau}^l] \mathbf{;} SVD(\mathbf{R}_\tau^l)= \mathbf{U}_{\tau}^l \mathbf{\Sigma} (\mathbf{V}_{\tau}^l)^T
\end{split}
\end{equation}
The threshold hyperparameter $\epsilon_{th}$ determines the number of basis vectors chosen from $\mathbf{U}_{\tau}^l$ to represent important gradient directions for any particular task. These vectors span a subspace in the gradient space which we define as the Core Gradient Space (CGS). They are added to the GPM matrix $\mathcal{M}= \{(\mathbf{M}^l)^L_{l=1}\}$, and the following update rule is used to obtain orthogonal gradient update $\tilde{\mathbf{g}}^i$ for the later tasks:
\vspace{-2mm}
\begin{equation} \label{orth}
    \tilde{\mathbf{g}}^i= \mathbf{g}^i - ({\mathbf{M}^l}{(\mathbf{M}^l)}^T)\mathbf{g}^i
\end{equation}
Here, $\mathbf{g}^i$ is the original local gradient update at agent $i$ at layer $l$, and the projection of $\mathbf{g}^i$ on CGS is $({\mathbf{M}^l}{\mathbf{M}^l}^T)\mathbf{g}^i$. Let the input space for a layer be of dimension $n_l$. This implies that $\mathbf{U}_{\tau}^l$ contains $n_l$ orthonormal basis vectors. Now based on $\epsilon_{th}$, after every task, a set of $r_l$ basis vectors corresponding to the top $r_l$ singular values are stored in $\mathcal{M}= \{(\mathbf{M}^l)^L_{l=1}\}$. Hence, $\tilde{\mathbf{g}}^i$ lies in a  $(n_l-r_l)$ dimensional orthogonal subspace denoted as the Residual Gradient Space (RGS).
The orthonormal basis vectors which span RGS are the remaining $(n_l-r_l)$ vectors contained in $\mathbf{U}_{\tau}^l$. We store them in the RGS Matrix $\mathcal{O}=\{(\mathbf{O}^l)^L_{l=1}\}$. Note that $n_l-r_l\hspace{-0.8mm}< n_l$, and $r_l$ increases as the task sequence progresses. We note that the gradient updates tend to lie in a lower dimensional subspace (i.e. RGS) whose dimensionality decreases based on $\epsilon_{th}$ and the number of tasks. 

In algorithm \ref{apx_alg:CoDeC}, model updates $\mathbf{q}^i_k$ are computed at every training iteration $k$. Since all the local gradients $\tilde{\mathbf{g}}^i_k$ lie in RGS, the updates $\mathbf{q}^i_k$ also lie in RGS. Therefore, we express layer-wise $\mathbf{q}^i_k$ as a linear combination of the basis vectors in ${\mathbf{O}^l}$ and find the associated coefficients $\mathbf{c}^i_k$. The neighbors of agent $i$ reconstruct the updates $\mathbf{q}^i_k$ from the received $\mathbf{c}^i_k$.
This encoding and decoding of $\mathbf{q}^i_k$ requires two additional matrix multiplications, as shown in lines 10 and 13 in algorithm \ref{apx_alg:CoDeC}. Our approach ensures that all agents have the same $\mathcal{M}$ and $\mathcal{O}$ matrices so that the reconstruction is exact. Hence, we achieve lossless communication compression by the virtue of taking orthogonal gradient updates to avoid catastrophic forgetting. \footnote{The PyTorch implementation of CoDeC can be found at \url{https://github.com/Sakshi09Ch/CoDeC}}

\section{Convergence Rate Analysis} \label{convrate}
In this section, we provide a convergence analysis for our algorithm. In particular, we provide an upper bound for $\|\nabla \mathcal{F}\left(\bar{\mathbf{x}}_{k}\right)\|^{2}$ , where $\nabla\mathcal{F}\left(\bar{\mathbf{x}}_{k}\right)$ is the average gradient achieved by the averaged model across all agents. Since our claims are valid for each task $\tau\in \{1,..,T\}$, the task subscript is dropped for the following analysis.
We make the following assumptions:
\vskip 2mm
\hspace{-4mm}\textbf{Assumption 1 - Lipschitz Gradients:} Each function $f_i(\mathbf{x})$ is L-smooth.
\vskip 2mm
\hspace{-4mm}\textbf{Assumption 2 - Bounded Variance:} The variance of the stochastic gradients is assumed to be bounded. There exist constants $\sigma$ and $\delta$ such that
\begin{equation}
\label{eq:inner_variance}
\mathbb{E}_{ d \sim \mathcal{D}_i} || \nabla F_i(\mathbf{x}; d) - \nabla f_i(\mathbf{x})||^2 \leq \sigma^2
\end{equation}
\begin{equation}
\label{eq:outer_variance}
\frac{1}{N} \sum_{i=1}^N || \nabla f_i(\mathbf{x}) - \nabla \mathcal{F}(\mathbf{x})||^2 \leq \delta^2 \hspace*{2mm} \forall i,x
\end{equation}
\textbf{Assumption 3 - Doubly Stochastic Mixing Matrix:} The mixing matrix $\mathbf{W}$ is a real doubly stochastic matrix with $\lambda_1(\mathbf{W})=1$ and
\vspace{-0.5em}
\begin{equation}
\label{eq:eigen}
max{\{|\lambda_2(\mathbf{W})|, |\lambda_N(\mathbf{W})|\}} \leq \sqrt{\rho}<1
\end{equation}
where $\lambda_i(\mathbf{W})$ is the $i^{th}$ largest eigenvalue of $\mathbf{W}$ and $\rho$ is a constant.

The above assumptions are commonly used in most decentralized learning works~\cite{DPSGD,ds,cga}. Since we modify the original gradient update $\mathbf{g}^i$, we introduce an additional assumption:
\newline \hspace{-5mm} \textbf{Assumption 4 - Bounded Orthogonal Updates:} 
For all agents $i$, we have:
\vspace{-1em}
\begin{equation}
\label{eq:orth}
   \|\tilde{\mathbf{g}}^i\| \leq \mu \|\mathbf{g}^i\|
   \vspace{-0.5em}
\end{equation}
where $\mu\in(0,1]$ signifies how constrained the gradient space is. In particular, $\mu$ encapsulates the average impact of the dimension of RGS subspace during training.

To ensure that the gradient update after projection is in the descent direction, we provide the following lemma:
\begin{lemma} \label{eq:descent} Given the original gradient update -$\mathbf{g}^i$ is in the descent direction, the orthogonal gradient update -$\tilde{\mathbf{g}}^i$ is also in the descent direction.
\end{lemma}
Before delving into the convergence analysis of CoDeC, we formally define the average consensus error as:
\vspace{-0.7em}
\begin{equation}
CE= \frac{1}{N} \sum_{i=1}^{N}\|\mathbf{\bar{x}}_{k}-\mathbf{x}^i_{k}\|^2 \hspace{2mm}\forall k\geq 0
\vspace{-0.5em}
\end{equation}
Here, $\mathbf{\bar{x}}_{k}$ represents the global average of the model parameters $\mathbf{x}^i_{k}$ at any given iteration $k$. CE is a measure of the effectiveness of gossip averaging in the decentralized learning scenario. In particular, a lower CE implies that the agents are closer to achieving a global consensus. We present the following lemma to bound the consensus error. 
\begin{lemma} \label{eq:lemma1} For all $K \geq 1$, we have:
\begin{equation}
\begin{split}
\sum_{k=0}^{K-1}\frac{1}{N}\sum_{i=1}^N\mathbb{E}\bigg[\bigg\|\bar{\mathbf{x}}_k-\mathbf{x}_k^i\bigg\|^2\bigg]\leq \frac{\eta^2\mu^2(3\sigma^2+3\delta^2)}{(1-\sqrt{\rho})^2}K\\
        &\hspace{-60mm}+\frac{3\eta^2 \mu^2}{(1-\sqrt{\rho})}\sum_{k=0}^{K-1}\mathbb{E}[\|\frac{1}{N}\sum_{i=1}^N\nabla f_i(\mathbf{x}_k^i)\|^2].
\end{split}
\end{equation}
\end{lemma}
For the proof of lemma \ref{eq:descent} and \ref{eq:lemma1}, please refer to the Appendix 
\ref{descentproof} and \ref{lem_2_proof} respectively. 

Theorem~\ref{theorem_1} presents the convergence of CoDeC (proof detailed in Appendix
~\ref{apx:theorem_1}).

\begin{theorem}\label{theorem_1} Given assumptions 1-4, let step size $\eta$ satisfy the following condition:
\begin{equation}
\label{eq:eta}
\begin{split}
\frac{1}{L} < \eta \leq \frac{\sqrt{(1-\sqrt{\rho})^2+12\mu^2}-(1-\sqrt{\rho})}{6L\mu^2}
\end{split}
\end{equation}
For all $K \geq 1$, we have
\begin{equation} \label{eq:theorem}
    \begin{split}
&\frac{1}{K}\sum_{k=0}^{K-1} \mathbb{E}\left[\left\|\nabla \mathcal{F}\left(\bar{\mathbf{x}}_{k}\right)\right\|^{2}\right] \leq
\frac{1}{C_{1}K}\bigg(\mathbb{E}\left[\mathcal{F}\left(\bar{\mathbf{x}}_{0}\right)-\mathcal{F}^*\right]\bigg) +\\
&C_2 \:\frac{\sigma^2}{N} + C_3\:\eta^2\mu^2\bigg(\frac{3 \sigma^2}{(1-\sqrt{\rho})^2}+\frac{3 \delta^2}{(1-\sqrt{\rho})^2}\bigg)
    \end{split}
\end{equation}
\vspace{-2mm}
where $C_1 = \frac{1}{2} (\eta-\frac{1}{L})$, $C_{2}= L\eta^{2} / 2C_{1}$,
$C_{3}= L^2 \eta/ 2C_{1}$. 

\end{theorem}
The result of theorem~\ref{theorem_1} shows that the norm of the average gradient achieved by the consensus model is upper-bounded by the suboptimality gap $(\mathcal{F}\left(\bar{\mathbf{x}}_{0}\right)-\mathcal{F}^*)$, the sampling variance ($\sigma$), the gradient variations ($\delta$), and the constraint on the gradient space ($\mu$). The suboptimality gap signifies how good the model initialization is.
$\sigma$ indicates the variation in gradients due to stochasticity, while $\delta$ is related to gradient variations across the agents. From equation \ref{eq:theorem}, we observe that $\mu$ appears in the last term and effectively scales $\sigma$ and $\delta$. 
A detailed explanation of the constraints on step size $\eta$ is presented in the Appendix \ref{apx:stepsize}. We present a corollary to show the convergence rate of CoDeC in terms of the training iterations. Note that we denote $a_n = O(b_n)$ if $a_n \leq cb_n$, where $c>0$ is a constant.

\begin{corollary}
\label{corol}
Suppose that the step size satisfies $\eta=O\Big(\sqrt{\frac{N}{K}}\Big)$.  
For a sufficiently large $K$ and some constant $C > 0$, 
\begin{align}
\frac{1}{K}\sum_{k=0}^{K-1} \mathbb{E}\left[\left\|\nabla \mathcal{F}\left(\bar{\mathbf{x}}_{k}\right)\right\|^{2}\right] \leq C\Bigg(\frac{1}{\sqrt{NK}}+\frac{1}{K}\Bigg)
\end{align}
\end{corollary}

The proof for Corollary~\ref{corol} is detailed in Appendix \ref{coro_1_proof}. It indicates that CoDeC achieves a convergence rate of $O(\frac{1}{\sqrt{NK}})$ for each task. This rate is similar to the well-known best result in decentralized SGD algorithms~\cite{DPSGD}. Since $\mu^2$ appears only in the higher order term $\frac{1}{K}$, it does not affect the order of the convergence rate.

\newcolumntype{g}{>{\columncolor{Gray}}c}
\begin{table*}[!ht]
\begin{center}
\begin{tabular}{cccgggggggg}
\hline
\multicolumn{1}{c}{\multirow{2}{*}{Dataset}} &
\multicolumn{1}{c}{\multirow{2}{*}{Agents}} &
\multicolumn{1}{c}{\multirow{2}{*}{Setup}} &
\multicolumn{3}{c}{Directed Ring} &
\multicolumn{1}{c}{} &
\multicolumn{3}{c}{Torus}
\\ 
\cline{4-6} \cline{8-10}

\multicolumn{1}{c}{} & & & \cellcolor{white} ACC(\%) & \cellcolor{white} BWT(\%) & \cellcolor{white} CC & \cellcolor{white}& \cellcolor{white} ACC(\%) & \cellcolor{white} BWT(\%) & \cellcolor{white} CC
\\
\hline 
\hline 
\rowcolor{white}
 &  & STL & 69.22 \scriptsize $\pm$ 0.10 & -  & - & & - & - & -\\
\rowcolor{white}
& & D-EWC & 53.12 \scriptsize $\pm$  0.62 & 0.24 \scriptsize $\pm$ 0.18 & 1x & & - & - & - \\
\rowcolor{white}
& & CoDeC(full comm.) & 57.54 \scriptsize $\pm$ 0.25 & -1.22 \scriptsize $\pm$ 0.22 & 1x & & - & - & - \\
 & \multirow{-4}{*}{4} & \cellcolor{Gray} CoDeC & 57.83 \scriptsize $\pm$ 0.25 & -0.95 \scriptsize $\pm$ 0.05 & 1.86x & & - & - & -\\
 
\hhline{~---------}
\rowcolor{white}
 &  & STL & 64.99 \scriptsize $\pm$ 0.41 & -  & - & & 65.17 \scriptsize $\pm$ 0.44 & - & -\\
\rowcolor{white}
&  & D-EWC & 50.52 \scriptsize $\pm$ 0.58 & 0.51 \scriptsize $\pm$ 0.09 & 1x & & 49.41 \scriptsize $\pm$ 0.88 & 0.29 \scriptsize $\pm$ 0.27 & 1x\\
\rowcolor{white}
& & CoDeC(full comm.) & 53.57 \scriptsize $\pm$ 0.38 & -0.65 \scriptsize $\pm$ 0.52 & 1x & & 53.54 \scriptsize $\pm$ 0.35 & -1.15 \scriptsize $\pm$ 0.41 & 1x\\
& \multirow{-4}{*}{8} & \cellcolor{Gray} CoDeC & 53.63 \scriptsize $\pm$ 0.25 & -0.43 \scriptsize $\pm$ 0.33 & 1.85x & & 53.62 \scriptsize $\pm$ 0.29 & -0.64 \scriptsize $\pm$ 0.36 & 1.86x\\

 \hhline{~---------}
 \rowcolor{white}
  &  & STL & 58.31 \scriptsize $\pm$ 0.49 & -  & - & & 59.29 \scriptsize $\pm$ 0.12 & - & -\\
  \rowcolor{white}
&  & D-EWC & 45.52 \scriptsize $\pm$ 0.60 & 0.22 \scriptsize $\pm$ 0.34 & 1x & & 44.53 \scriptsize $\pm$ 0.77 & -0.20 \scriptsize $\pm$ 0.56 & 1x\\
\rowcolor{white}
 & & CoDeC(full comm.) & 48.05 \scriptsize $\pm$ 0.45 & -0.38 \scriptsize $\pm$ 0.12 & 1x & & 48.19 \scriptsize $\pm$ 0.27 & -0.29 \scriptsize $\pm$ 0.11 & 1x\\
\multirow{-8}{*}{\rotatebox[origin=c]{90}{\hspace{14mm} Split CIFAR-100}}& \multirow{-4}{*}{16} & \cellcolor{Gray} CoDeC & 48.16 \scriptsize $\pm$ 0.33 & -0.18 \scriptsize $\pm$ 0.28 & 1.84x & & 48.36 \scriptsize $\pm$ 0.04 & -0.26 \scriptsize $\pm$ 0.31 & 1.84x\\
\hhline{----------}
 \rowcolor{white}
  &  & STL & 69.36 \scriptsize $\pm$ 0.78 & -  & - & & - & - & -\\
   \rowcolor{white}
& & D-EWC & 52.81 \scriptsize $\pm$  2.80 & -1.07 \scriptsize $\pm$ 2.03 & 1x & & - & - & - \\
 \rowcolor{white}
& & CoDeC(full comm.) & 60.03 \scriptsize $\pm$ 0.75 & 0.36 \scriptsize $\pm$ 1.01 & 1x & & - & - & - \\
& \multirow{-4}{*}{4}& \cellcolor{Gray} CoDeC & 59.00 \scriptsize $\pm$ 2.56 & -0.79 \scriptsize $\pm$ 0.27 & 1.51x & & - & - & -\\

 \hhline{~---------}
  \rowcolor{white}
  &  & STL & 63.13 \scriptsize $\pm$ 0.86 & -  & - & & 66.27 \scriptsize $\pm$ 1.47 & - & -\\
   \rowcolor{white}
&  & D-EWC & 46.39 \scriptsize $\pm$ 1.54 & -1.64 \scriptsize $\pm$ 1.11 & 1x & & 48.23 \scriptsize $\pm$ 3.14 & -1.02 \scriptsize $\pm$ 1.16 & 1x\\
 \rowcolor{white}
 & & CoDeC(full comm.) & 53.22 \scriptsize $\pm$ 1.82 & 0.08 \scriptsize $\pm$ 0.45 & 1x & & 59.90 \scriptsize $\pm$ 0.48 & 0.37 \scriptsize $\pm$ 0.24 & 1x\\
& \multirow{-4}{*}{8} & \cellcolor{Gray} CoDeC & 53.30 \scriptsize $\pm$ 1.25 & -0.46 \scriptsize $\pm$ 0.48 & 1.37x & & 59.97 \scriptsize $\pm$ 0.87 & -0.19 \scriptsize $\pm$ 0.98 & 1.53x\\

 \hhline{~---------}
  \rowcolor{white}
  &  & STL & 57.09 \scriptsize $\pm$ 1.55 & - & -  & & 63.51 \scriptsize $\pm$ 0.61 & - & -\\
   \rowcolor{white}
&  & D-EWC & 39.67 \scriptsize $\pm$ 1.37 & -1.32 \scriptsize $\pm$ 1.18 & 1x & & 45.14 \scriptsize $\pm$ 0.18 & -0.64 \scriptsize $\pm$ 0.23 & 1x\\
 \rowcolor{white}
 & & CoDeC(full comm.) & 45.29 \scriptsize $\pm$ 3.58 & -0.99 \scriptsize $\pm$ 1.40 & 1x & & 51.03 \scriptsize $\pm$ 2.51 & -0.01 \scriptsize $\pm$ 0.67 & 1x\\
\multirow{-10}{*}{\rotatebox[origin=c]{90}{\hspace{14mm} Split miniImageNet}}& \multirow{-4}{*}{16} & \cellcolor{Gray} CoDeC & 45.68 \scriptsize $\pm$ 0.77 & 0.61 \scriptsize $\pm$ 0.79 & 1.42x & & 51.32 \scriptsize $\pm$ 1.05 & 0.26 \scriptsize $\pm$ 0.56 & 1.39x\\
\hhline{----------}
  \rowcolor{white}
 &  & STL & 92.51 \scriptsize $\pm$ 0.18 & -  & - & & - & - & -\\
  \rowcolor{white}
& & D-EWC & 86.82 \scriptsize $\pm$  0.25 & -3.37 \scriptsize $\pm$ 0.80 & 1x & & - & - & - \\
 \rowcolor{white}
& & CoDeC(full comm.) & 87.24 \scriptsize $\pm$ 0.23 & -4.05 \scriptsize $\pm$ 0.05 & 1x & & - & - & - \\
& \multirow{-4}{*}{4} & \cellcolor{Gray} CoDeC & 87.41 \scriptsize $\pm$ 0.44 & -4.03 \scriptsize $\pm$ 0.30 & 2.13x & & - & - & -\\
\hhline{~---------}
 \rowcolor{white}
&  & STL & 92.31 \scriptsize $\pm$ 0.06 & -  & - & & 92.32 \scriptsize $\pm$ 0.15 & - & -\\
 \rowcolor{white}
&  & D-EWC & 85.69 \scriptsize $\pm$ 0.19 & -0.92 \scriptsize $\pm$ 0.14 & 1x & & 82.99 \scriptsize $\pm$ 3.25 & -2.10 \scriptsize $\pm$ 1.60 & 1x\\
 \rowcolor{white}
 & & CoDeC(full comm.) & 86.54 \scriptsize $\pm$ 0.04 & -4.37 \scriptsize $\pm$ 0.17 & 1x & & 85.92 \scriptsize $\pm$ 0.18 & -5.10 \scriptsize $\pm$ 0.17 & 1x\\
& \multirow{-4}{*}{8} & \cellcolor{Gray} CoDeC & 86.23 \scriptsize $\pm$ 0.22 & -4.61 \scriptsize $\pm$ 0.32 & 2.17x & & 86.15 \scriptsize $\pm$ 0.17 & -4.85 \scriptsize $\pm$ 0.26 & 2.19x\\
\hhline{~---------}
 \rowcolor{white}
 &  & STL & 92.16 \scriptsize $\pm$ 0.16 & -  & - & & 91.76 \scriptsize $\pm$ 0.09 & - & -\\
  \rowcolor{white}
&  & D-EWC & 82.19 \scriptsize $\pm$ 0.45 & -0.18 \scriptsize $\pm$ 0.05 & 1x & & 81.48 \scriptsize $\pm$ 0.12 & -0.56 \scriptsize $\pm$ 0.14 & 1x\\
 \rowcolor{white}
 & & CoDeC(full comm.) & 86.36 \scriptsize $\pm$ 0.15 & -4.36 \scriptsize $\pm$ 0.19 & 1x & & 84.91 \scriptsize $\pm$ 0.20 & -5.48 \scriptsize $\pm$ 0.22 & 1x\\
 \multirow{-8}{*}{\rotatebox[origin=c]{90}{\hspace{14mm} 5-Datasets}}& \multirow{-4}{*}{16} & \cellcolor{Gray} CoDeC & 86.41 \scriptsize $\pm$ 0.16 &-4.37 \scriptsize $\pm$ 0.24 & 2.16x & & 85.00 \scriptsize $\pm$ 0.55 & -5.52 \scriptsize $\pm$ 0.35 & 2.23x\\

 \hline
\end{tabular}
\end{center}
\vspace{-3mm}
\caption{Split CIFAR-100 over Alexnet, Split miniImageNet and 5-Datasets over ResNet-18 using directed ring and torus topology. STL is not a continual learning baseline, and serves as an upper bound for accuracy.}
\label{tab:allresults}
\vspace{-2mm}
\end{table*}

\section{Experimental Setup}
\noindent\textbf{Implementation details:} For each task, the data is independently and identically distributed (IID) across agents. 
The agents communicate model updates to their neighbors after every mini-batch update. We perform experiments across different graph topologies and sizes: directed ring with $N=4/8/16$ agents and undirected torus with $N=8/16$ agents. We evaluate CoDeC on three well-known continual learning benchmark datasets: 10-Split CIFAR-100 \cite{cifar}, 20-Split MiniImageNet \cite{miniI} and a sequence of 5-Datasets \cite{5D}. 10-Split CIFAR-100 is constructed by splitting CIFAR-100 into 10 tasks, where each task comprises of 10 classes. We use a 5-layer AlexNet for experiments with Split CIFAR-100. 20-Split miniImageNet has 20 sequential tasks, where each task comprises 5 classes. The sequence of 5-Datasets includes CIFAR-10, MNIST, SVHN \cite{svhn}, notMNIST \cite{nmnist} and Fashion MNIST\cite{fmnist}, where classification on each dataset is an individual task. For Split miniImageNet and 5-Datasets, we use a reduced ResNet18 architecture similar to \cite{gem}. The selection of threshold $(\epsilon_{th})$ for GPM and Fisher multiplier $(\lambda)$ for EWC is inspired by \cite{gpm}. In all our experiments, batch normalization parameters are learned for the first task and frozen for all subsequent tasks. We use `multi-head' setting, where each task has a separate final classifier with no constraints on gradient updates during training. Please refer to Appendix \ref{archdetails}, \ref{datasetdetails}, \ref{hyperdetails} for details related to  architectures, dataset statistics, and training hyperparameters, respectively.

\textbf{Baselines:} We implement D-EWC to establish a baseline, which extends Elastic Weight Consolidation (EWC) \cite{ewc} to a decentralized setting. EWC is one of the widely used regularization based continual learning approach that considers sequential task learning. This technique computes Fisher information matrix at the end of each task to constraint parameter updates for future tasks. Details on the implementation can be found in Appendix  \ref{baseline}. We also add a single task learning (STL) baseline, where all the tasks are learned sequentially in a decentralized setup without any constraints. This is equivalent to training a separate model in a decentralized manner for each task and will serve as an upper bound on the final average accuracy. STL is not a continual learning technique and may not be feasible in resource-constrained environments as it requires an excessive number of model parameters.

\textbf{Performance Metrics:} To evaluate the performance, we mainly focus on the following metrics: 
\vspace{-3.5mm}
\begin{itemize}
\setlength\itemsep{-0.4em}
\item \textit{Average Accuracy (ACC)}: This represents the average test classification accuracy of all the tasks. 
\item \textit{Backward Transfer (BWT)}: We measure the amount of forgetting by reporting backward transfer, which indicates the impact on the past knowledge after learning new tasks. A negative BWT is an indicator of catastrophic forgetting. 
\item \textit{Communication Compression (CC)}: We measure communication compression as the relative reduction in the communication cost achieved through our lossless compression scheme with respect to the full communication baseline. The agents communicate full precision (i.e. 32 bits) updates with their neighbors.
\end{itemize}
\vspace{-2.2mm}
ACC and BWT can be formally defined as:
\vspace{-0.5em}
\begin{figure*}[h]
    \centering
    \subfloat[\centering Split MiniImageNet]{{\includegraphics[width=0.45\textwidth]{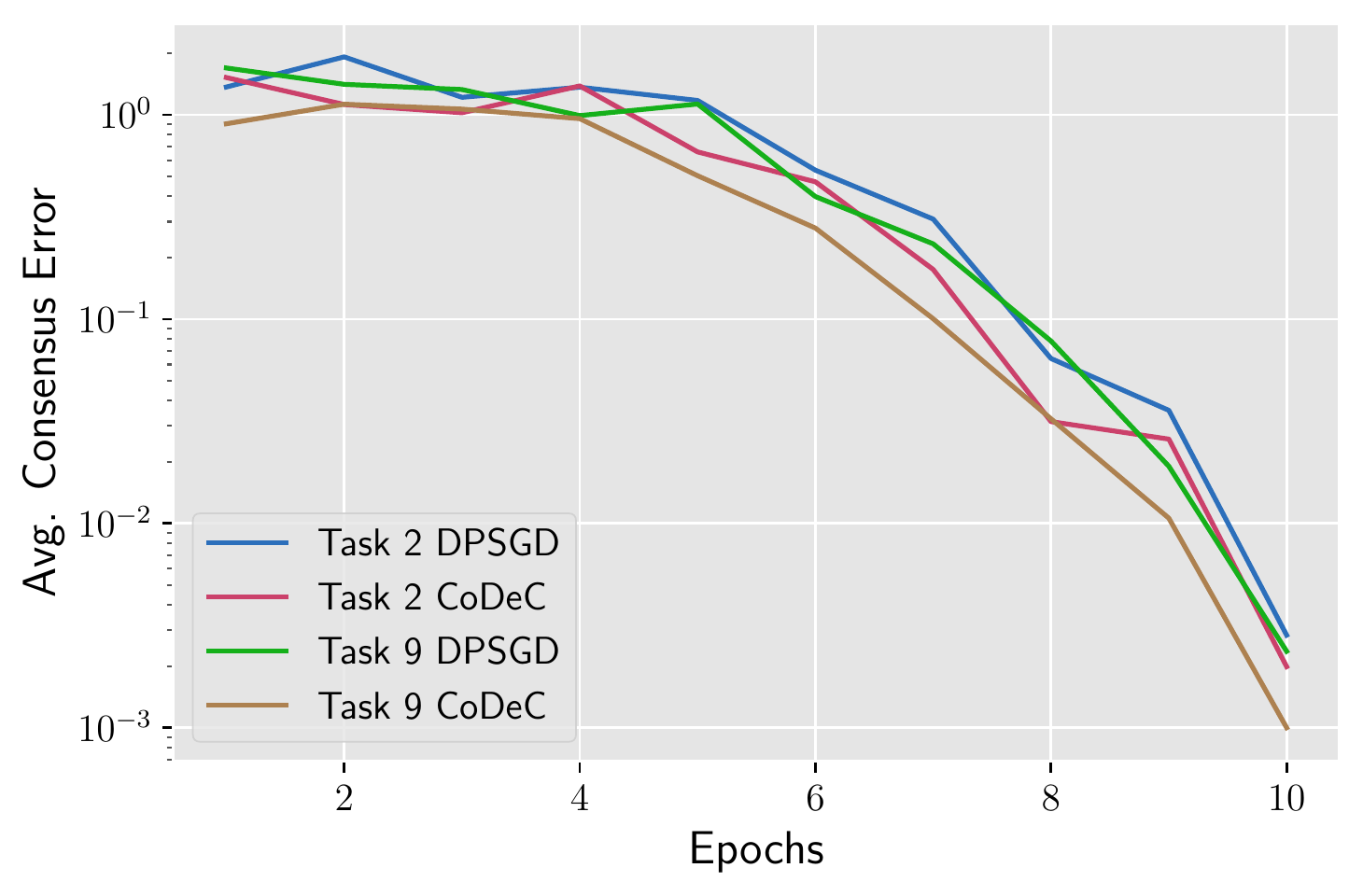} }}%
    \qquad
    \subfloat[\centering Split CIFAR-100]{{\includegraphics[width=0.45\textwidth]{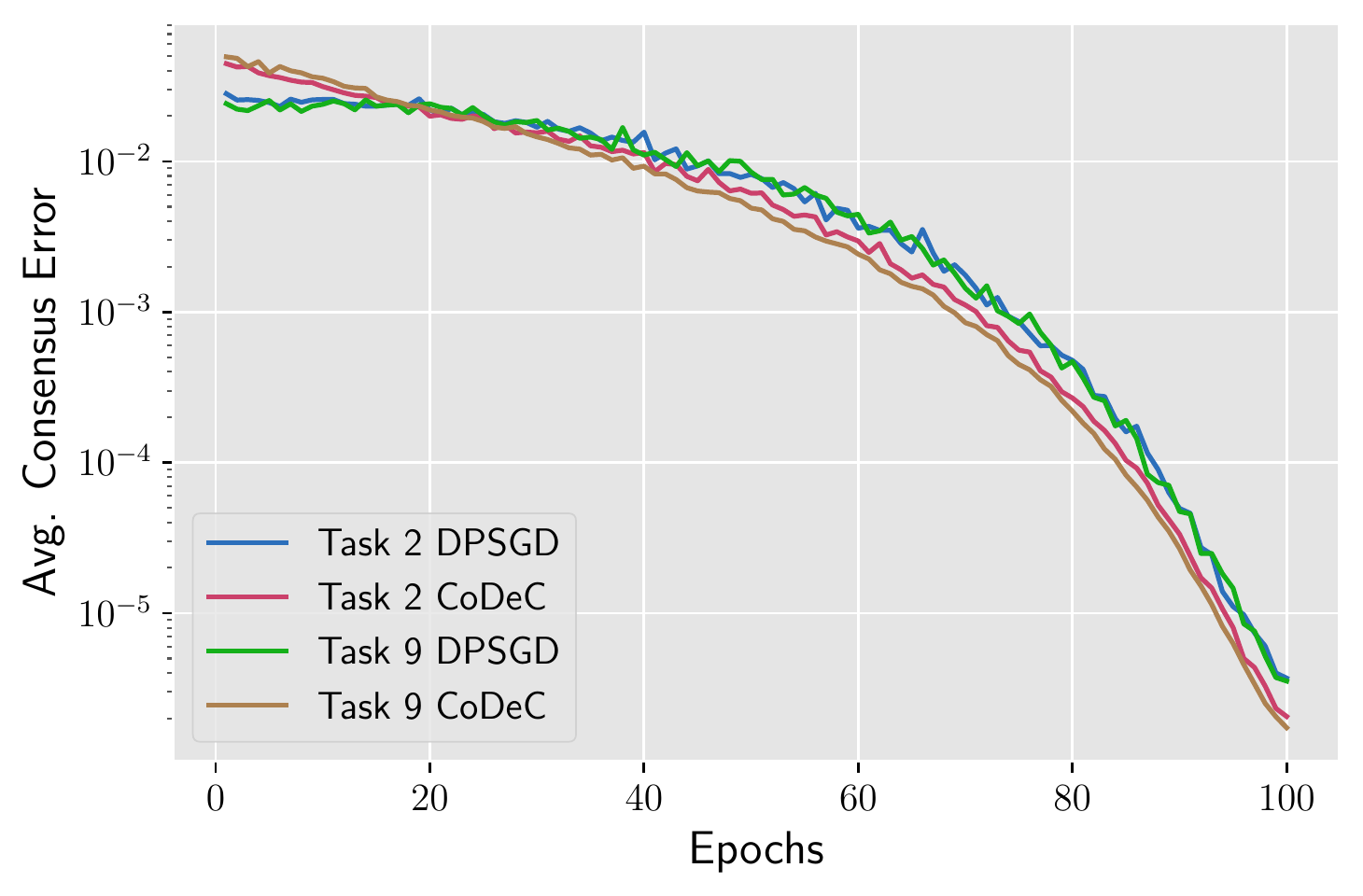} }}%
    \vspace{-3mm}
    \caption{Average consensus error for (a) Split MiniImageNet and (b) Split CIFAR-100 over a directed ring with 8 agents. Task `$\tau$' DPSGD (CoDeC) denotes consensus error when $\tau^{th}$ task is learned without (with) orthogonal gradient constraints.}%
    \label{fig:consen_er}
\end{figure*}
\begin{equation}\label{eq:sgp_acc}
    \text{ACC} = \frac{1}{T}\sum_{i=1}^T A_{T,i};\text{BWT} = \frac{1}{T-1}\sum_{i=1}^{T-1} A_{T,i} - A_{i,i} 
\end{equation}
Here, T is the total number of tasks and $A_{T,i}$ is the accuracy of the model on $i^{th}$ task after learning T tasks sequentially.

\section{Results and Discussions}

\subsection{Performance Analysis}
\begin{figure}[t!]
\centering
  \includegraphics[width=0.45\textwidth]{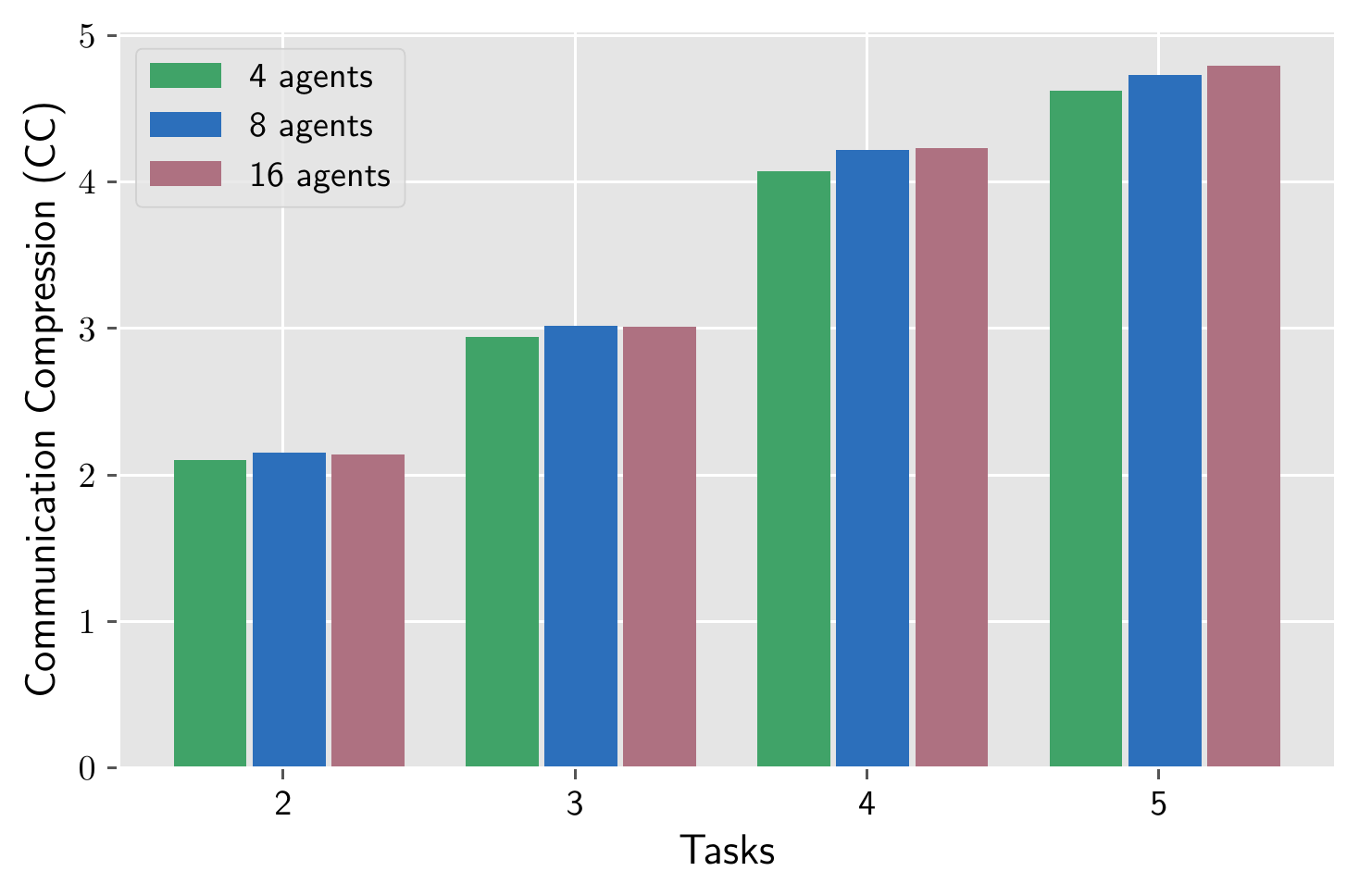}
  \vspace{-3mm}
  \caption{Task-wise CC for 5-Datasets over ResNet-18 with ring topology}
  \label{fig:taskCC}
\end{figure}

\vspace{-1.4mm}
We report results for Split CIFAR-100, Split MiniImageNet, and 5-Datasets across directed ring and torus graph topologies with different graph sizes in table \ref{tab:allresults}. In the directed ring topology, each agent has only 1 neighbor. Meanwhile, the torus topology has higher connectivity, with 3 and 4 neighbors for graph sizes of 8 and 16 agents respectively. 
We present two versions of our approach: CoDeC, which uses the lossless compression scheme and CoDeC(full comm.), an implementation with no communication compression. For Split CIFAR-100, we obtain $3-4\%$ better ACC than D-EWC with a similar order of BWT. Our proposed compression technique results in a 1.86x reduction in the communication cost on an average without any degradation in performance. For a longer task sequence Split MiniImageNet, we outperform D-EWC by $6-11\%$ in terms of ACC with marginally better BWT in some cases. 
We achieve 1.45x reduction in communication cost on average over a range of graph sizes and topologies. Results on 5-Datasets demonstrate learning across diverse datasets. As shown in table \ref{tab:allresults}, although we report better BWT for D-EWC, we achieve $0.5-4\%$ better accuracy with 2.2x reduced communication cost. In all our experiments, we observe that ACC decreases as we increase the graph size, while BWT remains of the similar order. 
Additionally, we present training times for CoDeC(full comm.), CoDeC and D-EWC in Appendix \ref{traintimesection}.
\newline The reduction in communication cost is a reflection of the constraints on the direction of gradient updates.
As the gradient updates are not constrained for the first task, they occupy the entire gradient space. However, gradient updates after learning task 1 are constrained to the RGS subspace, whose dimensionality decreases as the task sequence progresses. This implies an increase in compression ratios, which is clearly reflected in our results highlighting task-wise CC in figure \ref{fig:taskCC}.
In essence, as the gradient space becomes more constrained, it suffices for agents to communicate less with their neighbors. Hence, we achieve a CC of 2.1x for task 2, with this increasing up to 4.8x for task 5.

\subsection{Consensus Error}
We also investigate the effect of taking orthogonal gradient updates upon the average consensus error, which we formally define in section \ref{convrate}. 
In figure \ref{fig:consen_er}, we show CE with and without orthogonal updates. Figures \ref{fig:consen_er}(a) and \ref{fig:consen_er}(b) show the consensus error for task 2 and 9 after each training epoch for Split miniImageNet and Split CIFAR-100 respectively. As the training progresses, CE consistently reduces as expected. We observe that the rate of achieving consensus is similar for the two cases. In other words, CoDeC enables decentralized continual learning without hindering the gossip averaging mechanism. 

\section{Conclusion}
This work proposes CoDeC, a novel communication-efficient decentralized continual learning algorithm. CoDeC enables serverless training with spatially and temporally distributed private data and mitigates catastrophic forgetting by taking gradient steps orthogonal to the gradient directions important for previous tasks. These orthogonal gradient updates, and hence the model updates, lie in a lower \mbox{dimensional} gradient subspace. 
We exploit this fact to achieve lossless communication compression without requiring any additional hyperparameters. 
Further, we \mbox{provide theoretical} insights into the consensus error and the convergence rate of our algorithm. Our results demonstrate that CoDeC is very effective in learning distributed continual tasks with minimal backward transfer and up to 4.8x reduced communication overhead during training. 

\section{Acknowledgments}
This work was supported in part by, Center for Brain-inspired Computing (C-BRIC), a DARPA sponsored JUMP center, Semiconductor Research Corporation (SRC), National Science Foundation, the DoD Vannevar Bush Fellowship, and DARPA ShELL.

{\small
\bibliographystyle{ieee_fullname}
\bibliography{egbib}
}

\onecolumn
\appendix
\section{Appendix}
Proofs for the lemmas, theorems and corollaries presented in the main paper are detailed in \ref{descentproof}, \ref{lem_2_proof}, \ref{apx:theorem_1}, \ref{apx:stepsize} and \ref{coro_1_proof} sections. Details related to the network architectures and datasets used in our experiments are presented in \ref{archdetails} and \ref{datasetdetails} respectively. We list all our training hyperparameters in \ref{hyperdetails}. We also provide details about implementation of our baseline D-EWC in \ref{baseline}. Training times are reported in \ref{traintimesection}. Some additional results related to task-wise compression and lossless nature of our proposed compression scheme are available in \ref{addresults}. 

\subsection{Proof of Lemma \ref{eq:descent}}\label{descentproof}
The orthogonal projection $\tilde{\mathbf{g}}^i$ of the original gradient update $\mathbf{g}^i$ with respect to GPM is obtained as:
\begin{equation} \label{orth2}
    \tilde{\mathbf{g}}^i= \mathbf{g}^i - ({\mathbf{M}^l}{(\mathbf{M}^l)}^T)\mathbf{g}^i
\end{equation}

From the above equation we can write:
\begin{equation} \label{gi}
    \mathbf{g}^i= \tilde{\mathbf{g}}^i + ({\mathbf{M}^l}{(\mathbf{M}^l)}^T)\mathbf{g}^i 
\end{equation}

We have:
\begin{equation} \label{dotprod}
    \langle\mathbf{g}^i, \tilde{\mathbf{g}}^i\rangle = \langle\tilde{\mathbf{g}}^i + ({\mathbf{M}^l}{(\mathbf{M}^l)}^T)\mathbf{g}^i, \tilde{\mathbf{g}}^i\rangle = \langle\tilde{\mathbf{g}}^i, \tilde{\mathbf{g}}^i\rangle+\langle({\mathbf{M}^l}{(\mathbf{M}^l)}^T)\mathbf{g}^i, \tilde{\mathbf{g}}^i\rangle
\end{equation}

Since $\tilde{\mathbf{g}}^i$ and  ${\mathbf{M}^l}{(\mathbf{M}^l)}^T)\mathbf{g}^i$ are orthogonal to each other:
\begin{equation} \label{ortho_prop}
 \langle({\mathbf{M}^l}{(\mathbf{M}^l)}^T)\mathbf{g}^i, \tilde{\mathbf{g}}^i\rangle = 0
\end{equation}

Substituting equation~\ref{ortho_prop} into equation~\ref{dotprod}:
\begin{equation}\label{fn_lemma1}
    \langle\mathbf{g}^i, \tilde{\mathbf{g}}^i\rangle = \langle\tilde{\mathbf{g}}^i, \tilde{\mathbf{g}}^i\rangle = \|\tilde{\mathbf{g}}^i\|^2 \geq 0
\end{equation}

From the above equation, we see that the dot product is greater than or equal to $0$. This implies that if -$\mathbf{g}^i$ is in the descent direction, -$\tilde{\mathbf{g}}^i$ is also in the descent direction.

\subsection{Proof of Lemma \ref{eq:lemma1}}\label{lem_2_proof}
This section presents the detailed proof for Lemma 4.2. We follow the same approach as \cite{cga}. 
The update rule for our algorithm is as follows: 
\begin{equation}\label{update}
\begin{split}
    &\mathbf{\bar{x}}_k=\mathbf{\bar{x}}_{k-1}-\eta\frac{1}{N}\sum_{i=1}^N\tilde{\mathbf{g}}^i_{k-1}\\
\end{split}
\end{equation}
$\mathbf{\bar{x}}_k$ denotes the averaged model across all the agents at a given iteration $k$. For the rest of the analysis, the initial value will be directly set to $0$. From equation \ref{update} we have:

\begin{equation}\label{update2}
\begin{split}
    &\mathbf{\bar{x}}_{k+1}-\mathbf{\bar{x}}_{k}= -\eta\frac{1}{N}\sum_{i=1}^N\tilde{\mathbf{g}}^i_{k}\\
\end{split}
\end{equation}
We introduce some key notations and properties:
\begin{equation}\label{notation}
    \begin{split}
        &\mathbf{Q}=\frac{1}{N}\mathbf{1}\mathbf{1}^\top\\
        &\tilde{\mathbf{G}}_k\triangleq [\tilde{\mathbf{g}}^1_k,\tilde{\mathbf{g}}^2_k,...,\tilde{\mathbf{g}}^N_k]\\
        &\mathbf{X}_k\triangleq[\mathbf{x}^1_k,\mathbf{x}^2_k,...,\mathbf{x}^N_k]\\
        &\mathbf{G}_k\triangleq[\mathbf{g}^1_k,\mathbf{g}^2_k,...,\mathbf{g}^N_k]\\
        &\mathbf{H}_k\triangleq[\nabla f_1(\mathbf{x}^1_k),\nabla f_2(\mathbf{x}^2_k),...,\nabla f_N(\mathbf{x}^N_k)]\\
    \end{split}
\end{equation}
For all the above matrices, $\|\mathbf{A}\|_\mathfrak{F}^2=\sum_{i=1}^N\|\mathbf{a}_i\|^2$, where $\mathbf{a}_i$ is the $i$-th column of the matrix $\mathbf{A}$.  Thus, we obtain:
\begin{equation}
    \|\mathbf{X}_k(\mathbf{I}-\mathbf{Q})\|_\mathfrak{F}^2=\sum_{i=1}^N\|\mathbf{x}^i_k-\bar{\mathbf{x}}_k\|^2.
\end{equation}
For each doubly stochastic matrix $\mathbf{W}$, the following properties hold true
\begin{itemize}
    \item $\mathbf{Q}\mathbf{W}=\mathbf{W}\mathbf{Q}$;
    \item $(\mathbf{I}-\mathbf{Q})\mathbf{W}=\mathbf{W}(\mathbf{I}-\mathbf{Q})$;
    \item For any integer $k\geq 1$, $\|(\mathbf{I}-\mathbf{Q})\mathbf{W}\|_\mathfrak{S}\leq(\sqrt{\rho})^k$, where $\|\cdot\|_\mathfrak{S}$ is the spectrum norm of a matrix.
\end{itemize}
For $N$ arbitrary real square matrices $\mathbf{A}_i, i\in\{1,2,...,N\}$,
\begin{equation}\label{fact_1}
    \|\sum_{i=1}^N\mathbf{A}_i\|^2_\mathfrak{F}\leq \sum_{i=1}^N\sum_{j=1}^N\|\mathbf{A}_i\|_\mathfrak{F}\|\mathbf{A}_j\|_\mathfrak{F}.
\end{equation}

\par We are now ready to prove Lemma 4.2. Since $ \mathbf{X}_{k} =\mathbf{X}_{k-1}\mathbf{W} - \eta \tilde{\mathbf{G}}_k$ we have:

\begin{equation}
\begin{split}
    \mathbf{X}_{k}(\mathbf{I}-\mathbf{Q}) =\mathbf{X}_{k-1}(\mathbf{I}-\mathbf{Q})\mathbf{W} - \eta \tilde{\mathbf{G}}_k (\mathbf{I}-\mathbf{Q})
\end{split}
\end{equation}
Applying the above equation $k$ times we have: 

\begin{equation}
\begin{split}
    &\mathbf{X}_{k}(\mathbf{I}-\mathbf{Q}) =\mathbf{X}_{0}(\mathbf{I}-\mathbf{Q})\mathbf{W}^k -\sum_{\tau=1}^{k}\eta \tilde{\mathbf{G}}_{\tau}(\mathbf{I}-\mathbf{Q})\mathbf{W}^{k-\tau} = -\eta \sum_{\tau=1}^{k}\tilde{\mathbf{G}}_{\tau}(\mathbf{I}-\mathbf{Q})\mathbf{W}^{k-\tau}
\end{split}
\end{equation} 

\begin{equation}\label{expand}
\begin{split}
     &\mathbb{E}\bigg[\bigg\|\mathbf{X}_k(\mathbf{I}-\mathbf{Q})\bigg\|^2_\mathfrak{F}\bigg] = \eta^2 \underbrace{\mathbb{E}\bigg[\bigg\|\sum_{\tau=0}^{k-1}\tilde{\mathbf{G}}_{\tau}(\mathbf{I}-\mathbf{Q})\mathbf{W}^{k-1-\tau}\bigg\|^2_\mathfrak{F}\bigg]}_{I}
\end{split}
\end{equation}
We find the upper bound for term $I$.

\begin{equation}\label{I}
    \begin{split}
        & \mathbb{E}\bigg[\bigg\|\sum_{\tau=0}^{k-1}\tilde{\mathbf{G}}_{\tau}(\mathbf{I}-\mathbf{Q})\mathbf{W}^{k-1-\tau}\bigg\|^2_\mathfrak{F}\bigg] \overset{a}{\leq} 
        \sum_{\tau=0}^{k-1}\sum_{\tau^\prime=0}^{k-1}\mathbb{E}\bigg[\bigg\|\tilde{\mathbf{G}}_{\tau}(\mathbf{I}-\mathbf{Q})\mathbf{W}^{k-1-\tau}\bigg\|_\mathfrak{F}\bigg\|\tilde{\mathbf{G}}_{\tau^\prime}(\mathbf{I}-\mathbf{Q})\mathbf{W}^{k-1-\tau^\prime}\bigg\|_\mathfrak{F}\bigg]  \\
        &\leq \sum_{\tau=0}^{k-1}\sum_{\tau^\prime=0}^{k-1}\rho^{(k-1-\frac{\tau+\tau^\prime}{2})}\mathbb{E}[\|\tilde{\mathbf{G}}_{\tau}\|_\mathfrak{F}\|\tilde{\mathbf{G}}_{\tau^\prime}\|_\mathfrak{F}] \overset{b}{\leq}\sum_{\tau=0}^{k-1}\sum_{\tau^\prime=0}^{k-1}\mu^2\rho^{(k-1-\frac{\tau+\tau^\prime}{2})}\mathbb{E}[\|\mathbf{G}_{\tau}\|_\mathfrak{F}\|\mathbf{G}_{\tau^\prime}\|_\mathfrak{F}]\\
        &\overset{c}{\leq}\sum_{\tau=0}^{k-1}\sum_{\tau^\prime=0}^{k-1}\mu^2\rho^{(k-1-\frac{\tau+\tau^\prime}{2})}\bigg(\frac{1}{2}\mathbb{E}[\|\mathbf{G}_{\tau}\|_\mathfrak{F}^2]+\frac{1}{2}\mathbb{E}[\|\mathbf{G}_{\tau^\prime}\|_\mathfrak{F}^2]\bigg) \\
        &= \sum_{\tau=0}^{k-1}\sum_{\tau^\prime=0}^{k-1}\mu^2\rho^{(k-1-\frac{\tau+\tau^\prime}{2})}\mathbb{E}[\|\mathbf{G}_{\tau}\|_\mathfrak{F}^2]\overset{d}{\leq} \frac{\mu^2}{(1-\sqrt{\rho})}\sum_{\tau=0}^{k-1}\rho^{(\frac{k-1-\tau}{2})}\mathbb{E}[\|\mathbf{G}_{\tau}\|_\mathfrak{F}^2]
    \end{split}
\end{equation}

\noindent (a) follows from equation~\ref{fact_1}. 
\newline (b) follows from assumption 4.
\newline (c) follows from the inequality $xy \leq \frac{1}{2}(x^2+y^2)$ for any two real numbers $x,y$. 
\newline (d) is derived from $\sum_{\tau_1=0}^{k-1}\rho^{k-1-\frac{\tau_1+\tau}{2}} \leq \frac{\rho^{\frac{k-1-\tau}{2}}}{1-\sqrt{\rho}}$.
\vspace{6mm}
\newline \noindent We proceed with finding the bounds for $\mathbb{E}[\|\mathbf{G}_{\tau}\|_\mathfrak{F}^2]$:
\begin{equation}\label{E_G}
    \begin{split}
        &\mathbb{E}[\|\mathbf{G}_\tau\|^2_\mathfrak{F}] = \mathbb{E}[\|\mathbf{G}_\tau- \mathbf{H}_\tau+ \mathbf{H}_\tau-\mathbf{H}_\tau \mathbf{Q}+\mathbf{H}_\tau \mathbf{Q}\|^2_\mathfrak{F}]\\
        & \leq 3\mathbb{E}[\|\mathbf{G}_\tau- \mathbf{H}_\tau\|^2_\mathfrak{F}]+ 3\mathbb{E}[\|\mathbf{H}_\tau(I-\mathbf{Q})\|^2\mathfrak{F}]+3\mathbb{E}[\|\mathbf{H}_\tau \mathbf{Q}\|^2_\mathfrak{F}]
        \overset{a}{\leq} 
        3N\sigma^2+3N\delta^2+3 \mathbb{E}[\|\frac{1}{N}\sum_{i=1}^N\nabla f_i(\mathbf{x}^i_\tau)\|^2]
    \end{split}
\end{equation}
(a) holds because $\mathbb{E}[\|\mathbf{H}_\tau \mathbf{Q}\|^2_\mathfrak{F}]\leq\mathbb{E}[\|\frac{1}{N}\sum_{i=1}^N\nabla f_i(\mathbf{x}^i_\tau)\|^2]$
\newline Substituting (\ref{E_G}) in ~(\ref{I}):

\begin{equation}\label{prefinal}
    \begin{split}
        & \mathbb{E}\bigg[\bigg\|\sum_{\tau=0}^{k-1}\tilde{\mathbf{G}}_{\tau}(\mathbf{I}-\mathbf{Q})\mathbf{W}^{k-1-\tau}\bigg\|^2_\mathfrak{F}\bigg] \leq \frac{\mu^2}{(1-\sqrt{\rho})}\sum_{\tau=0}^{k-1}\rho^{(\frac{k-1-\tau}{2})}\bigg[3N\sigma^2+3N\delta^2+3 \mathbb{E}[\|\frac{1}{N}\sum_{i=1}^N\nabla f_i(\mathbf{x}^i_\tau)\|^2]\bigg]\\
        &\leq \frac{3N\mu^2(\sigma^2+\delta^2)}{(1-\sqrt{\rho})^2}+\frac{3N\mu^2}{(1-\sqrt{\rho})}\sum_{\tau=0}^{k-1}\rho^{(\frac{k-1-\tau}{2})}\mathbb{E}[\|\frac{1}{N}\sum_{i=1}^N\nabla f_i(\mathbf{x}^i_\tau)\|^2]
    \end{split}
\end{equation}
Substituting~(\ref{prefinal}) into the main inequality (\ref{expand}):

\begin{equation}\label{final1}
    \begin{split}
        &\mathbb{E}\bigg[\bigg\|\mathbf{X}_k(\mathbf{I}-\mathbf{Q})\bigg\|^2_\mathfrak{F}\bigg] \leq 
       \eta^2\mu^2\bigg(\frac{3N\sigma^2}{(1-\sqrt{\rho})^2}+\frac{3N\delta^2}{(1-\sqrt{\rho})^2}\bigg)+ \frac{3N\eta^2\mu^2}{(1-\sqrt{\rho})}\sum_{\tau=0}^{k-1}\rho^{(\frac{k-1-\tau}{2})}\mathbb{E}[\|\frac{1}{N}\sum_{i=1}^N\nabla f_i(\mathbf{x}^i_\tau)\|^2]
    \end{split}
\end{equation}
Summing over $k\in\{1,\dots, K-1\}$ and noting that $\mathbb{E}\bigg[\bigg\|\mathbf{X}_0(\mathbf{I}-\mathbf{Q})\bigg\|^2_\mathfrak{F}\bigg] = 0$:

\begin{equation}\label{final2}
    \begin{split}
        &\sum_{k=1}^{K-1}\mathbb{E}\bigg[\bigg\|\mathbf{X}_k(\mathbf{I}-\mathbf{Q})\bigg\|^2_\mathfrak{F}\bigg] \leq CK + \frac{3N\eta^2\mu^2}{(1-\sqrt{\rho})}\sum_{k=1}^{K-1}\sum_{\tau=0}^{k-1}\rho^{(\frac{k-1-\tau}{2})}\mathbb{E}[\|\frac{1}{N}\sum_{i=1}^N\nabla f_i(\mathbf{x}^i_\tau)\|^2]\leq \\
        & CK + \frac{3N\eta^2\mu^2}{(1-\sqrt{\rho})}\sum_{k=0}^{K-1}\frac{1-\rho^{(\frac{K-1-k}{2})}}{1-\sqrt{\rho}}\mathbb{E}[\|\frac{1}{N}\sum_{i=1}^N\nabla f_i(\mathbf{x}^i_k)\|^2]\leq CK + \frac{3N\eta^2\mu^2}{(1-\sqrt{\rho})}\sum_{k=0}^{K-1}\mathbb{E}[\|\frac{1}{N}\sum_{i=1}^N\nabla f_i(\mathbf{x}^i_k)\|^2]\\
        &where\  C = \eta^2\mu^2\bigg(\frac{3N\sigma^2+3N\delta^2}{(1-\sqrt{\rho})^2}\bigg)\\
    \end{split}
\end{equation}
\newline Dividing both sides by $N$:

\begin{equation}\label{final2-2}
    \begin{split}
        \sum_{k=1}^{K-1}\frac{1}{N}\mathbb{E}\bigg[\bigg\|\mathbf{X}_k(\mathbf{I}-\mathbf{Q})\bigg\|^2_\mathfrak{F}\bigg] \leq
        &\eta^2\mu^2\bigg(\frac{3 \sigma^2}{(1-\sqrt{\rho})^2}+\frac{3 \delta^2}{(1-\sqrt{\rho})^2}\bigg)K+ \frac{3 \eta^2\mu^2}{(1-\sqrt{\rho})}\sum_{k=0}^{K-1}\mathbb{E}[\|\frac{1}{N}\sum_{i=1}^N\nabla f_i(\mathbf{x}^i_k)\|^2]
    \end{split}
\end{equation}
This directly implies:

\begin{equation}
    \begin{split}\label{consenerror_bound}
        \sum_{k=0}^{K-1}\frac{1}{N}\sum_{i=1}^N\mathbb{E}\bigg[\bigg\|\bar{\mathbf{x}}_k-\mathbf{x}^i_k\bigg\|^2\bigg] \leq \eta^2\mu^2\bigg(\frac{3 \sigma^2}{(1-\sqrt{\rho})^2}+\frac{3 \delta^2}{(1-\sqrt{\rho})^2}\bigg)K+ \frac{3 \eta^2\mu^2}{(1-\sqrt{\rho})}\sum_{k=0}^{K-1}\mathbb{E}[\|\frac{1}{N}\sum_{i=1}^N\nabla f_i(\mathbf{x}^i_k)\|^2]
    \end{split}
\end{equation}

\vspace{5mm}
\subsection{Proof for Theorem \ref{theorem_1}}\label{apx:theorem_1}
When $\mathcal{F}$ is $L$-smooth, we have:
\begin{equation}\label{main0}
    \mathbb{E}[\mathcal{F}(\bar{\mathbf{x}}_{k+1})] \leq \mathbb{E}[\mathcal{F}(\bar{\mathbf{x}}_{k})]+\underbrace{\mathbb{E}[\langle\nabla\mathcal{F}(\bar{\mathbf{x}}_{k}),\bar{\mathbf{x}}_{k+1}- \bar{\mathbf{x}}_{k}\rangle]}_{I}+ \frac{L}{2} \mathbb{E}[\|\bar{\mathbf{x}}_{k+1} - \bar{\mathbf{x}}_{k}\|^2 ]
\end{equation}
We proceed by analysing $I$:
\begin{equation}\label{innerprod}
\begin{split}
\mathbb{E}[\langle\nabla\mathcal{F}(\bar{\mathbf{x}}_{k}),\bar{\mathbf{x}}_{k+1}- \bar{\mathbf{x}}_{k}\rangle]= \mathbb{E}[\langle\nabla\mathcal{F}(\bar{\mathbf{x}}_{k}), -\eta \bigg( \frac{1}{N} \sum_{i=1}^N\tilde{\mathbf{g}}^i_k\bigg)\rangle]
\end{split}
\end{equation}

\begin{equation}\label{helper2}
\begin{split}
 \mathbb{E}[\langle\nabla \mathcal{F}\left(\bar{\mathbf{x}}_{k}\right), -\eta\bigg(\frac{1}{N} \sum_{i=1}^{N} \tilde{\mathbf{g}}_{k}^{i}\bigg)\rangle]=  &\mathbb{E}[\langle\nabla \mathcal{F}\left(\bar{\mathbf{x}}_{k}\right), -\eta \bigg(\frac{1}{N} \sum_{i=1}^{N}\tilde{\mathbf{g}}_{k}^{i}-\mathbf{g}_{k}^{i}+\mathbf{g}_{k}^{i}\bigg)\rangle] \\
&\hspace{-3mm}=\underbrace{\mathbb{E}[\langle\nabla \mathcal{F}\left(\bar{\mathbf{x}}_{k}\right), -\eta \bigg(\frac{1}{N} \sum_{i=1}^{N}\tilde{\mathbf{g}}_{k}^{i}-\mathbf{g}_{k}^{i}\bigg)\rangle]}_{II}+\underbrace{
\mathbb{E}[\langle\nabla \mathcal{F}\left(\bar{\mathbf{x}}_{k}\right) , -\eta \bigg(\frac{1}{N} \sum_{i=1}^{N}\mathbf{g}_{k}^{i}\bigg)\rangle]}_{III}
\end{split}
\end{equation}
We first analyse $II$:
\begin{equation}\label{helper2-1}
    \begin{split}
        -\eta\mathbb{E}[\langle\nabla \mathcal{F}\left(\bar{\mathbf{x}}_{k}\right), \frac{1}{N} \sum_{i=1}^{N}\left(\tilde{\mathbf{g}}_{k}^{i}-\mathbf{g}_{k}^{i}\right)\rangle] \leq \frac{1}{2L}\mathbb{E}[\|\nabla \mathcal{F}(\bar{\mathbf{x}}_{k})\|^{2}]+\frac{L \eta^2}{2}\mathbb{E}[\|\frac{1}{N} \sum_{i=1}^{N} (\tilde{\mathbf{g}}_{k}^{i}- \mathbf{g}_k^i)\|^{2}]
    \end{split}
\end{equation}
This  holds as 
$\langle \mathbf{a},\mathbf{b} \rangle \leq \frac{1}{2}\|\mathbf{a}\|^2 + \frac{1}{2}\|\mathbf{b}\|^2$.
\newline Analysing $III$:
\begin{equation}\label{helper2-2}
    \begin{split}
        \mathbb{E}\bigg[\langle\nabla \mathcal{F}\left(\bar{\mathbf{x}}_{k}\right) , -\eta \bigg(\frac{1}{N} \sum_{i=1}^{N}\mathbf{g}_{k}^{i}\bigg)\rangle\bigg] =  -\eta\mathbb{E}\bigg[\langle\nabla\mathcal{F}(\bar{\mathbf{x}}_{k}),\frac{1}{N}\sum_{i=1}^{N}\nabla f_i(\mathbf{x}^i_k)\rangle\bigg]
    \end{split}
\end{equation}
With the aid of the equity $\langle \mathbf{a},\mathbf{b} \rangle = \frac{1}{2}[\|\mathbf{a}\|^2 + \|\mathbf{b}\|^2 - \|\mathbf{a}-\mathbf{b}\|^2]$, we have :

\begin{equation}\label{helper2-2-1}
\begin{split}
&\langle\nabla \mathcal{F}\left(\bar{\mathbf{x}}_{k}\right), \frac{1}{N} \sum_{i=1}^{N} \nabla f_{i}\left(\mathbf{x}_{k}^{i}\right)\rangle=\frac{1}{2}\left(\|\nabla \mathcal{F}\left(\bar{\mathbf{x}}_{k}\right)\|^{2}+\| \frac{1}{N} \sum_{i=1}^{N}
\nabla f_{i}(\mathbf{x}_{k}^{i})\|^{2}-\underbrace{\| \nabla \mathcal{F}(\bar{\mathbf{x}}_{k})-\frac{1}{N} \sum_{i=1}^{N} \nabla f_{i}(\mathbf{x}_{k}^{i}) \|^{2}}_\star\right) 
\end{split}
\end{equation}
Analysing $\star$:
\begin{equation}\label{star}
\begin{split}
&\|\nabla\mathcal{F}(\bar{\mathbf{x}}_{k})- \frac{1}{N}\sum_{i=1}^{N}\nabla f_i(\mathbf{x}^i_k)\|^2 = \|\frac{1}{N}\sum_{i=1}^{N}\nabla f_i(\bar{\mathbf{x}}_{k})- \frac{1}{N}\sum_{i=1}^{N}\nabla f_i(\mathbf{x}^i_k)\|^2 \\
&\leq \frac{1}{N}\sum_{i=1}^{N}\|\nabla f_i(\bar{\mathbf{x}}_{k})- \nabla f_i(\mathbf{x}^i_k)\|^2 \leq \frac{1}{N}\sum_{i=1}^{N} L^2 \|\bar{\mathbf{x}}_{k}- \mathbf{x}^i_k\|^2
\end{split}
\end{equation}
Substituting (\ref{star}) back into (\ref{helper2-2-1}), we have:
\begin{equation}\label{helper2-2-1_1}
\begin{split}
&\langle\nabla \mathcal{F}\left(\bar{\mathbf{x}}_{k}\right), \frac{1}{N} \sum_{i=1}^{N} \nabla f_{i}\left(\mathbf{x}_{k}^{i}\right)\rangle {\geq} \frac{1}{2}\left(\|\nabla \mathcal{F}(\bar{\mathbf{x}}_{k})\|^{2}+
\|\frac{1}{N} \sum_{i=1}^{N} \nabla f_{i}(\mathbf{x}_{k}^{i})\|^{2}-L^{2} \frac{1}{N} \sum_{i=1}^{N}\|\bar{\mathbf{x}}_{k}-\mathbf{x}_{k}^{i}\|^{2}\right)
\end{split}
\end{equation}
Substituting ~(\ref{helper2-1}) and ~(\ref{helper2-2-1_1}) into ~(\ref{helper2}), and ~(\ref{helper2}) into~(\ref{innerprod}):

\begin{equation}\label{main2}
\begin{split}
&\mathbb{E}[\langle\nabla\mathcal{F}(\bar{\mathbf{x}}_{k}),\bar{\mathbf{x}}_{k+1}- \bar{\mathbf{x}}_{k}\rangle] \leq \bigg(\frac{1}{2L}-\frac{\eta}{2}\bigg)\mathbb{E}[\|\nabla \mathcal{F}(\bar{\mathbf{x}}_{k})\|^{2}]+\frac{L \eta^2}{2}\mathbb{E}[\|\frac{1}{N} \sum_{i=1}^{N} (\tilde{\mathbf{g}}_{k}^{i}- \mathbf{g}_k^i)\|^{2}] \\
&\hspace{40mm}-\frac{\eta}{2}\left(\mathbb{E}[\|\frac{1}{N} \sum_{i=1}^{N} \nabla f_{i}(\mathbf{x}_{k}^{i})\|^{2}]-L^{2} \mathbb{E}[\frac{1}{N} \sum_{i=1}^{N}\|\bar{\mathbf{x}}_{k}-\mathbf{x}_{k}^{i}\|^{2}]\right)
\end{split}
\end{equation}

\noindent From equation (\ref{update2}), we have:
\begin{equation}\label{eqlem_3}
    \mathbb{E}[\|\bar{\mathbf{x}}_{k+1}-\bar{\mathbf{x}}_k\|^2]=\eta^2\mathbb{E}[\|\frac{1}{N}\sum_{i=1}^N\tilde{\mathbf{g}}^i_k\|^2].
\end{equation}

\noindent Substituting (\ref{main2}) and (\ref{eqlem_3}) in (\ref{main0}):

\begin{equation}\label{main3}
\begin{split}
    &\mathbb{E}[\mathcal{F}(\bar{\mathbf{x}}_{k+1})] \leq \mathbb{E}[\mathcal{F}(\bar{\mathbf{x}}_{k})]+ \bigg(\frac{1}{2L}-\frac{\eta}{2}\bigg)\mathbb{E}[\|\nabla \mathcal{F}(\bar{\mathbf{x}}_{k})\|^{2}]+\frac{L \eta^2}{2}\mathbb{E}[\|\frac{1}{N} \sum_{i=1}^{N} (\tilde{\mathbf{g}}_{k}^{i}- \mathbf{g}_k^i)\|^{2}] \\
&\hspace{20mm}-\frac{\eta}{2}\mathbb{E}[\|\frac{1}{N} \sum_{i=1}^{N} \nabla f_{i}(\mathbf{x}_{k}^{i})\|^{2}]+\frac{\eta L^2}{2} \mathbb{E}[\frac{1}{N} \sum_{i=1}^{N}\|\bar{\mathbf{x}}_{k}-\mathbf{x}_{k}^{i}\|^{2}]+\frac{\eta^2L}{2}\mathbb{E}[\|\frac{1}{N}\sum_{i=1}^N\tilde{\mathbf{g}}^i_k\|^2]
    \end{split}
\end{equation}

\noindent Rearranging the terms and dividing by $C_1 = \bigg(\frac{\eta}{2}-\frac{1}{2L}\bigg) > 0$ to find the bound for $\mathbb{E}[\|\nabla\mathcal{F}(\bar{\mathbf{x}}_{k})\|^2]$:

\begin{equation}\label{main4}
\begin{split}
    &\mathbb{E}[\|\nabla\mathcal{F}(\bar{\mathbf{x}}_{k})\|^2] \leq
    \frac{1}{C_1}\bigg(\mathbb{E}[\mathcal{F}(\bar{\mathbf{x}}_{k})]-\mathbb{E}[\mathcal{F}(\bar{\mathbf{x}}_{k+1})]\bigg)+
    C_2 \:\bigg(\underbrace{\mathbb{E}[\|\frac{1}{N} \sum_{i=1}^{N} (\tilde{\mathbf{g}}_{k}^{i}- \mathbf{g}_k^i)\|^{2}]+\mathbb{E}[\|\frac{1}{N}\sum_{i=1}^N\tilde{\mathbf{g}}^i_k\|^2]}_{\star}\bigg)\\
    &\hspace{25mm}+C_3\: \mathbb{E}[\frac{1}{N} \sum_{i=1}^{N}\|\bar{\mathbf{x}}_{k}-\mathbf{x}_{k}^{i}\|^{2}]- C_4\:\mathbb{E}[\|\frac{1}{N} \sum_{i=1}^{N} \nabla f_{i}(\mathbf{x}_{k}^{i})\|^{2}]\\
    &where\ C_{2}=L\eta^2/2C_1, C_{3}=L^2\eta/2C_1,C_4=\eta/2C_1.
    \end{split}
\end{equation}

\noindent We first analyze $\star$:
\begin{equation}\label{star_2}
\begin{split}
&\mathbb{E}[\|\frac{1}{N} \sum_{i=1}^{N} (\tilde{\mathbf{g}}_{k}^{i}- \mathbf{g}_k^i)\|^{2}]+\mathbb{E}[\|\frac{1}{N}\sum_{i=1}^N\tilde{\mathbf{g}}^i_k\|^2] = \frac{1}{N^2}\mathbb{E}[\| \sum_{i=1}^{N} (\tilde{\mathbf{g}}_{k}^{i}- \mathbf{g}_k^i)\|^{2}]+\|\sum_{i=1}^N\tilde{\mathbf{g}}^i_k\|^2]\\
&\overset{a}{=} \frac{1}{N^2}\mathbb{E}[\| \sum_{i=1}^{N} (\mathbf{M}\mathbf{M^T}\mathbf{g}_k^i)\|^{2}]+\|\sum_{i=1}^N((\mathbf{I}-\mathbf{M}\mathbf{M^T})\mathbf{g}^i_k)\|^2] \overset{b}{=} \frac{1}{N^2}\mathbb{E}[\| \mathbf{M}\mathbf{M^T} \sum_{i=1}^{N} (\mathbf{g}_k^i)\|^{2}]+\|(\mathbf{I}-\mathbf{M}\mathbf{M^T})\sum_{i=1}^N(\mathbf{g}^i_k)\|^2]\\
&=\mathbb{E}[\| \sum_{i=1}^{N}  \frac{1}{N} \mathbf{g}_k^i\|^{2}] \overset{c}{\leq} \bigg(\frac{\sigma^2}{N}+\mathbb{E} \bigg[\bigg\|\frac{1}{N}\sum_{i=1}^N\nabla f_i(\mathbf{x}^i)\bigg\|^2\bigg]\bigg)
\end{split}
\end{equation}

\noindent(a) follows from the fact that $\tilde{\mathbf{g}}_{k}^{i}$ is an orthogonal projection of $\mathbf{g}_{k}^{i}$, and it is defined by the GPM matrix $\mathbf{M}$. 

\noindent (b) follows from all agents having the same GPM matrix $\mathbf{M}$
 
\noindent (c) is the conclusion of Lemma $1$ in ~\cite{yu2019linear}.

\noindent Substituting (\ref{star_2}) into (\ref{main4}) and summing over $k \in \{0,1,\dots, K-1\}$: 

\begin{equation}
    \begin{split}
&\sum_{k=0}^{K-1} \mathbb{E}\left[\left\|\nabla \mathcal{F}\left(\bar{\mathbf{x}}_{k}\right)\right\|^{2}\right] \leq
\frac{1}{C_{1}}\bigg(\mathbb{E}\left[\mathcal{F}\left(\bar{\mathbf{x}}_{0}\right)-\mathcal{F}\left(\bar{\mathbf{x}}_{k}\right)\right]\bigg) + C_2 \:\sum_{k=0}^{K-1}\bigg(\frac{\sigma^2}{N}+\mathbb{E} \bigg[\bigg\|\frac{1}{N}\sum_{i=1}^N\nabla f_i(\mathbf{x}_k^i)\bigg\|^2\bigg]\bigg)\\
&+ C_3\: \sum_{k=0}^{K-1}\mathbb{E}\left[\frac{1}{N}\sum_{i=1}^{N}\left\|\bar{\mathbf{x}}_{k}-\mathbf{x}_{k}^{i}\right\|^{2}\right]- C_4\:\sum_{k=0}^{K-1}\mathbb{E}\left[\left\|\frac{1}{N} \sum_{i=1}^{N} \nabla f_{i}(\mathbf{x}_{k}^{i})\right\|^{2}\right]
    \end{split}
\end{equation}
Dividing both sides by $K$:
\begin{equation}
    \begin{split}
&\frac{1}{K}\sum_{k=0}^{K-1} \mathbb{E}\left[\left\|\nabla \mathcal{F}\left(\bar{\mathbf{x}}_{k}\right)\right\|^{2}\right] \leq
\frac{1}{C_{1}K}\bigg(\mathbb{E}\left[\mathcal{F}\left(\bar{\mathbf{x}}_{0}\right)-\mathcal{F}^*\right]\bigg) +  C_2 \:\frac{\sigma^2}{N} + C_2 \:\sum_{k=0}^{K-1}\frac{1}{K}\left(\mathbb{E} \left[\bigg\|\frac{1}{N}\sum_{i=1}^N\nabla f_i(\mathbf{x}_k^i)\bigg\|^2\right]\right)\\
&+ \frac{C_3}{K}\: \sum_{k=0}^{K-1}\mathbb{E}\left[\frac{1}{N}\sum_{i=1}^{N}\left\|\bar{\mathbf{x}}_{k}-\mathbf{x}_{k}^{i}\right\|^{2}\right]- C_4\:\sum_{k=0}^{K-1}\frac{1}{K}\mathbb{E}\left[\left\|\frac{1}{N} \sum_{i=1}^{N} \nabla f_{i}(\mathbf{x}_{k}^{i})\right\|^{2}\right]
    \end{split}
\end{equation}
Using Lemma \ref{eq:lemma1} in the above equation, we have:
\begin{equation}
    \begin{split}
&\frac{1}{K}\sum_{k=0}^{K-1} \mathbb{E}\left[\left\|\nabla \mathcal{F}\left(\bar{\mathbf{x}}_{k}\right)\right\|^{2}\right] \leq
\frac{1}{C_{1}K}\bigg(\mathbb{E}\left[\mathcal{F}\left(\bar{\mathbf{x}}_{0}\right)-\mathcal{F}^*\right]\bigg) +  C_2 \:\frac{\sigma^2}{N} + C_2 \:\sum_{k=0}^{K-1}\frac{1}{K}\left(\mathbb{E} \left[\bigg\|\frac{1}{N}\sum_{i=1}^N\nabla f_i(\mathbf{x}_k^i)\bigg\|^2\right]\right)\\
&+ \frac{C_3}{K}\: \left[\eta^2\mu^2\bigg(\frac{3 \sigma^2}{(1-\sqrt{\rho})^2}+\frac{3 \delta^2}{(1-\sqrt{\rho})^2}\bigg)K+ \frac{3 \eta^2\mu^2}{(1-\sqrt{\rho})}\sum_{k=0}^{K-1}\mathbb{E}[\|\frac{1}{N}\sum_{i=1}^N\nabla f_i(\mathbf{x}^i_k)\|^2]\right]\\
&- C_4\:\sum_{k=0}^{K-1}\frac{1}{K}\mathbb{E}\left[\left\|\frac{1}{N} \sum_{i=1}^{N} \nabla f_{i}(\mathbf{x}_{k}^{i})\right\|^{2}\right]
    \end{split}
\end{equation}

\noindent Rearranging the terms:
\begin{equation}
    \begin{split}
&\frac{1}{K}\sum_{k=0}^{K-1} \mathbb{E}\left[\left\|\nabla \mathcal{F}\left(\bar{\mathbf{x}}_{k}\right)\right\|^{2}\right] \leq
\frac{1}{C_{1}K}\bigg(\mathbb{E}\left[\mathcal{F}\left(\bar{\mathbf{x}}_{0}\right)-\mathcal{F}^*\right]\bigg) +  C_2 \:\frac{\sigma^2}{N} + C_3\:\eta^2\mu^2\bigg(\frac{3 \sigma^2}{(1-\sqrt{\rho})^2}+\frac{3 \delta^2}{(1-\sqrt{\rho})^2}\bigg) \\
&\hspace{35mm}+ \left(C_2+\frac{3C_3\eta^2\mu^2}{(1-\sqrt{\rho})}-C_4\right) \:\left(\frac{1}{K}\sum_{k=0}^{K-1}\mathbb{E} \left[\bigg\|\frac{1}{N}\sum_{i=1}^N\nabla f_i(\mathbf{x}_k^i)\bigg\|^2\right]\right)\\
    \end{split}
\end{equation}

\noindent When $\left(C_2+\frac{3C_3\eta^2\mu^2}{(1-\sqrt{\rho})}-C_4\right) \leq 0$, we have:
\begin{equation}\label{upperbound_final}
    \begin{split}
&\frac{1}{K}\sum_{k=0}^{K-1} \mathbb{E}\left[\left\|\nabla \mathcal{F}\left(\bar{\mathbf{x}}_{k}\right)\right\|^{2}\right] \leq
\frac{1}{C_{1}K}\bigg(\mathbb{E}\left[\mathcal{F}\left(\bar{\mathbf{x}}_{0}\right)-\mathcal{F}^*\right]\bigg) +  C_2 \:\frac{\sigma^2}{N} + C_3\:\eta^2\mu^2\bigg(\frac{3 \sigma^2}{(1-\sqrt{\rho})^2}+\frac{3 \delta^2}{(1-\sqrt{\rho})^2}\bigg)
    \end{split}
\end{equation}

\subsection{Discussion on the Step Size}\label{apx:stepsize}
Recall the condition $C1>0$. This implies $\eta > \frac{1}{L}$.
\newline The condition for equation (\ref{upperbound_final}) to be true is $\left(C_2+\frac{3C_3\eta^2\mu^2}{(1-\sqrt{\rho})}-C_4\right) \leq 0$.
Therefore, we have:
\begin{equation}
\begin{split}
\frac{3L^2\eta^2\mu^2}{(1-\sqrt{\rho})}+\eta L-1 \leq 0
\end{split}
\end{equation}
Solving this inequality, combining the fact that $\eta > 0$, we have then the specific form of $\eta^{*}$:
\begin{equation}
\begin{split}
\eta^* = \frac{\sqrt{(1-\sqrt{\rho})^2+12\mu^2}-(1-\sqrt{\rho})}{6L\mu^2}
\end{split}
\end{equation}
Hence, the step size $\eta$ is defined as
\begin{equation}
\begin{split}
\frac{1}{L} < \eta \leq \frac{\sqrt{(1-\sqrt{\rho})^2+12\mu^2}-(1-\sqrt{\rho})}{6L\mu^2}
\end{split}
\end{equation}
\subsection{Proof for Corollary \ref{corol}}\label{coro_1_proof}
According to equation (\ref{upperbound_final}), on the right hand side, there are three terms with different coefficients with respect to the step size $\eta$. We separately investigate each term:
\newline $\eta=\mathcal{O}\bigg(\sqrt{\frac{N}{K}}\bigg)$ implies  $C_1 = \mathcal{O}\bigg(\sqrt{\frac{N}{K}}\bigg)$. Therefore for the first term:
\begin{equation}
\begin{split}
\frac{\mathcal{F}(\bar{\mathbf{x}}_0)-\mathcal{F}^*}{C_1K}=\mathcal{O}\bigg(\frac{1}{\sqrt{NK}}\bigg)
\end{split}
\end{equation}
For the second term:
\begin{equation}
\begin{split}
\frac{C_2}{N}=\mathcal{O}\bigg(\frac{1}{N}\sqrt{\frac{N}{K}}\bigg)= \mathcal{O}\bigg(\frac{1}{\sqrt{NK}}\bigg)
\end{split}
\end{equation}
For the third term:
\begin{equation}
\begin{split}
\eta^2C_3=\mathcal{O}\bigg(\frac{N}{K}\bigg)
\end{split}
\end{equation}
By omitting $N$ in non-dominant terms, there exists a constant $C>0$ such that the overall convergence rate is as follows:
\begin{equation}
    \frac{1}{K}\sum_{k=0}^{K-1} \mathbb{E}\left[\left\|\nabla \mathcal{F}\left(\bar{\mathbf{x}}_{k}\right)\right\|^{2}\right] \leq C\Bigg(\frac{1}{\sqrt{NK}}+\frac{1}{K}\Bigg), 
\end{equation}
which suggests when $N$ is fixed and $K$ is sufficiently large, CoDeC enables the convergence rate of $O(\frac{1}{\sqrt{NK}})$.

\subsection{Network Architecture} \label{archdetails}
\begin{itemize}
    \item \textbf{AlexNet-like architecture}: For our experiments, we scale the output channels in each layer of the architecture used in \cite{hat}. The network consists of 3 convolutional layers of 16, 32, and 64 filters with 4 $\times$ 4, 3 $\times$ 3, and 2 $\times$ 2 kernel sizes, respectively and 2 fully connected layers of 512 units each. A 2 $\times$ 2 max-pooling layer follows the convolutional layers. Rectified linear units are used as activations. Dropout of 0.2 is used for the first two layers and 0.5 for the rest of the layers.
    \vspace{-2mm}
    \item \textbf{Reduced ResNet18 architecture}: This is similar to the architecture used by \cite{gem}. We replace the 4 $\times$ 4 average-pooling layer with a 2 $\times$ 2 layer. For experiments with miniImageNet, we use convolution with stride 2 in the first layer. 
\end{itemize}
All the networks use ReLU in the hidden units and softmax with cross entropy loss in the final layer.

\subsection{Datasets}\label{datasetdetails}
Table \ref{tab_app1} and \ref{tab_app2} provide the details related to the datasets used in our experiments. The training samples/tasks are independently and identically distributed (IID) across agents without any data overlap. For instance, for a graph size of 4 agents, each agent has $5000/4=1250$ training samples for a particular task in Split CIFAR-100.
\renewcommand{\arraystretch}{1.1}
\begin{table}[h]\centering
\begin{tabular}{lccccc}\hline
& \multicolumn{1}{c}{\textbf{Split CIFAR-100}} & \phantom{a}& \multicolumn{1}{c}{\textbf{Split  miniImageNet}}\\
\hline
num. of tasks    &   10  && 20   \\
input size   & $3\times 32\times 32$ && $3\times 84\times 84$\\
\# Classes/task  & 10 && 5  \\
\# Training samples/tasks  &5,000&& 2,500\\
\# Test samples/tasks & 1,000&& 500\\
\hline
\end{tabular}
\vspace{-2mm}
\caption{Dataset Statistics for Split CIFAR-100 and Split-miniImageNet}
\label{tab_app1}
\end{table}

\renewcommand{\arraystretch}{1.1}
\begin{table}[h]\centering
\begin{tabular}{lcccccccc}\hline
& \multicolumn{1}{c}{\textbf{CIFAR-10}} & \multicolumn{1}{c}{\textbf{MNIST}} & \multicolumn{1}{c}{\textbf{SVHN}} & \multicolumn{1}{c}{\textbf{Fashion MNIST}} & \multicolumn{1}{c}{\textbf{notMNIST}}\\
\hline
Classes    & 10 & 10 & 10 & 10 & 10 \\
\# Training samples/tasks  & 50,000 & 60,000 & 73,257 & 60,000 & 16,853\\
\# Test samples/tasks & 10,000 & 10,000 & 26,032 & 10,000 & 1,873\\
\hline
\end{tabular}
\vspace{-2mm}
\caption{5-Datasets Statistics}
\label{tab_app2}
\end{table}
\vspace{-3mm}
\subsection{Hyperparameters} \label{hyperdetails}
All our experiments were run for three randomly chosen seeds. We decay the learning rate by a factor of 10 after 50\% and 75\% of the training, unless mentioned otherwise.
\vspace{1mm}
\newline \textbf{Hyperparameters for Split CIFAR-100 on AlexNet:} For CoDeC, we use an initial learning rate of 0.01. $\epsilon_{th}$ is initially set to 0.97 and incremented by 0.003 for each task. For D-EWC, we use an initial learning rate of 0.05, and $\lambda$ is set to 5000. We use a mini-batch size of 22 per agent, and we run all our experiments for a total of 100 epochs for each task.
\vspace{1mm}
\newline \textbf{Hyperparameters for Split miniImageNet on ResNet-18:} For CoDeC, we use an initial learning rate of 0.1. $\epsilon_{th}$ is initially set to 0.985 and incremented by 0.0003 for each task. For D-EWC, we use an initial learning rate of 0.03, and $\lambda$ is set to 5000. We use a mini-batch size of 10 per agent. All our experiments are run for a total of 10 epochs for each task.
\vspace{1mm}
\newline \textbf{Hyperparameters for 5-Datasets on ResNet-18:} For CoDeC, we use an initial learning rate of 0.1. $\epsilon_{th}$ is set to 0.965 for each task. For D-EWC, we use an initial learning rate of 0.03, and $\lambda$ is set to 5000. We use a mini-batch size of 32 per agent, and we run all our experiments for a total of 50 epochs for each task.
\vspace{1mm}
\par The values of threshold $\epsilon_{th}$, Fisher multiplier $\lambda$ and learning rate are inspired by GPM\cite{gpm}. The average consensus error plots shown in figure \ref{fig:consen_er} were obtained with a cosine annealing based learning rate scheduling instead of the step decay mentioned earlier.
\subsection{Baseline Implementation} \label{baseline}
Algorithm \ref{apx_alg:dewc} demonstrates the flow of D-EWC, the baseline which extends EWC\cite{ewc} to a decentralized setting.
\begin{algorithm}[ht]
\textbf{Input:} Each agent $i \in [1,N]$ initializes model parameters $\mathbf{x}_0^{i}$, step size $\eta$, mixing matrix $\mathbf{W}=[w_{ij}]_{i,j \in [1,N]}$, $\hat{\mathbf{x}}_{(0)}^i=0$, $\mathbf{F}^l= [\hspace{1mm}]$ for all layers $l= 1,2,...L$, Fisher Matrix $\mathcal{F}^i= \{(\mathbf{F}^l)^L_{l=1}\}$, old model parameters $\mathbf{x}^i_{(0)}=0$, $\mathcal{N}(i)$:~neighbors of agent $i$ (including itself), $T$: total tasks, $K$: number of training iterations\\

Each agent simultaneously implements the 
T\text{\scriptsize RAIN}( ) procedure\\
\text{\small 1.}  \textbf{procedure} T\text{\scriptsize RAIN}( ) \\
\text{\small 2.}  \hspace{4mm}\textbf{for} $\tau=1,\hdots,T$ \textbf{do}\\
\text{\small 3.}  \hspace{8mm}\textbf{for} $k=0,1,\hdots,K-1$ \textbf{do}\\
\text{\small 4.}  \hspace*{12mm}$d_{\tau, i} \sim \mathcal{D}_{\tau, i}$\\
\text{\small 5.}  \hspace*{12mm}$\tilde{f}_{\tau,i}(d_{\tau, i};\mathbf{x}^i_{k})=f_{\tau,i}(d_{\tau, i}; \mathbf{x}^i_{k})+ \sum_{l=0}^L \frac{\lambda}{2}\mathbf{F}^l(\mathbf{x}^{i,l}_{k}-\mathbf{x}^{i,l}_{\tau-1}) $\\
\text{\small 6.}  \hspace*{12mm}$\mathbf{g}^{i}_{k}=\nabla \tilde{f}_{\tau,i}(d_{\tau, i}; \mathbf{x}^i_{k}) $\\
\text{\small 7.}  \hspace*{12mm}$\mathbf{x}_{(k+\frac{1}{2})}^i=\mathbf{x}_k^{i}- \eta \mathbf{g}^i_{k} $\\
\text{\small 8.}  \hspace*{12mm}$\mathbf{x}_{k+1}^{i}=\mathbf{x}_{(k+\frac{1}{2})}^i+ \sum_{j \in \mathcal{N}(i)} w_{ij}(\hat{\mathbf{x}}_k^{j}-\mathbf{x}_k^{i})$\\    
\text{\small 9.} \hspace*{12mm}$\mathbf{q}_{k}^i= \mathbf{x}^i_{k+1}-\mathbf{x}^i_{k}$\\
\text{\small 10.} \hspace*{10mm} for each $j \in \mathcal{N}(i)$ \textbf{do}\\
\text{\small 11.}    \hspace*{15mm} Send $\mathbf{q}_{k}^i$ and receive $\mathbf{q}_{k}^j$\\
\text{\small 12.}     \hspace*{15mm} $\hat{\mathbf{x}}^j_{(k+1)}= \mathbf{q}_{k}^j+\hat{\mathbf{x}}_k^{j}$\\
\text{\small 13.}    \hspace*{11mm}\textbf{end}\\   
\text{\small 14.} \hspace*{8mm}\textbf{end}\\
\text{\small 15.} \hspace*{8mm}Save $\mathbf{x}^i_{\tau}$\\
\text{\small 16.}    \hspace*{8mm} \# EWC Update  \\
\text{\small 17.}   \hspace*{8mm}Update $\mathbf{F}^l$ for each layer $l$\\
\text{\small 18.}   \hspace*{8mm}Update $\mathcal{F}^i=\{(\mathbf{F}^l)^L_{l=1}\}$\\
\text{\small 19.}    \hspace*{8mm} $p= random(1,2,...N)$ \\
\text{\small 20.}    \hspace*{8mm} \textbf{if} $i==p$ \textbf{do} \\
\text{\small 21.}     \hspace*{12mm}Gather $\mathcal{F}^i$ from all agents\\
\text{\small 22.}     \hspace*{12mm}$\mathcal{F}= avg(\mathcal{F}^1, \mathcal{F}^2, ....\mathcal{F}^N)$\\
\text{\small 23.}     \hspace*{12mm}Send $\mathcal{F}$ to all agents\\
\text{\small 24.}    \hspace*{8mm} \textbf{end}\\
\text{\small 25.}   \hspace*{4mm}\textbf{end}\\
\text{\small 26.} \textbf{return}
\caption{Decentralized Elastic Weight Consolidation (\textit{D-EWC})}
\label{apx_alg:dewc}
\end{algorithm}
The loss function minimized in EWC is of the form $\tilde{f}_{\tau,i}(d^i_{\tau, k};\mathbf{x}^i_{k})$ shown in line 5, algorithm \ref{apx_alg:dewc}. Here, $\lambda$ is a regularization coefficient which signifies the importance given to the past tasks. $\mathbf{x}^{i,l}_{k}$ and $\mathbf{x}^{i,l}_{\tau-1}$ represent model parameters for a particular layer $l$.
Unlike CoDeC, here we generate the Fisher matrix $\mathcal{F}^i$ at each agent and then do a global averaging step before utilizing it for continually learning the next task. We do so because EWC\cite{ewc} utilizes the entire training data to generate the Fisher matrix.

\subsection{Training Time} \label{traintimesection}
\begin{table}[h]
\begin{center}
\begin{tabular}{ccc} 
\hline
\textbf{Dataset} & \textbf{Setup} & \textbf{Training time}\\
\hline
\hline
\multirow{3}{*}{Split CIFAR-100} & CoDeC(full comm.) & 1 \\
& CoDeC & 1.36 \\
& D-EWC & 1.58 \\
\hline
\multirow{3}{*}{Split miniImageNet} & CoDeC(full comm.) & 1 \\
& CoDeC & 1.21 \\
& D-EWC & 0.91 \\
\hline
\multirow{3}{*}{5-Datasets} & CoDeC(full comm.) & 1 \\
& CoDeC & 1.48 \\
& D-EWC & 1.47 \\
\hline
\end{tabular}
\end{center}
\vspace{-3mm}
\caption{Training time for Split CIFAR-100, Split miniImageNet and 5-Datasets over a directed ring topology with 8 agents}\label{traintime}
\vspace{-4mm}
\end{table}
The training times are presented in table \ref{traintime} and normalized with respect to the runtime of CoDeC(full comm.).

\subsection{Additional Results}\label{addresults}
\begin{figure}[h!]
\centering
  \includegraphics[width=0.6\textwidth]{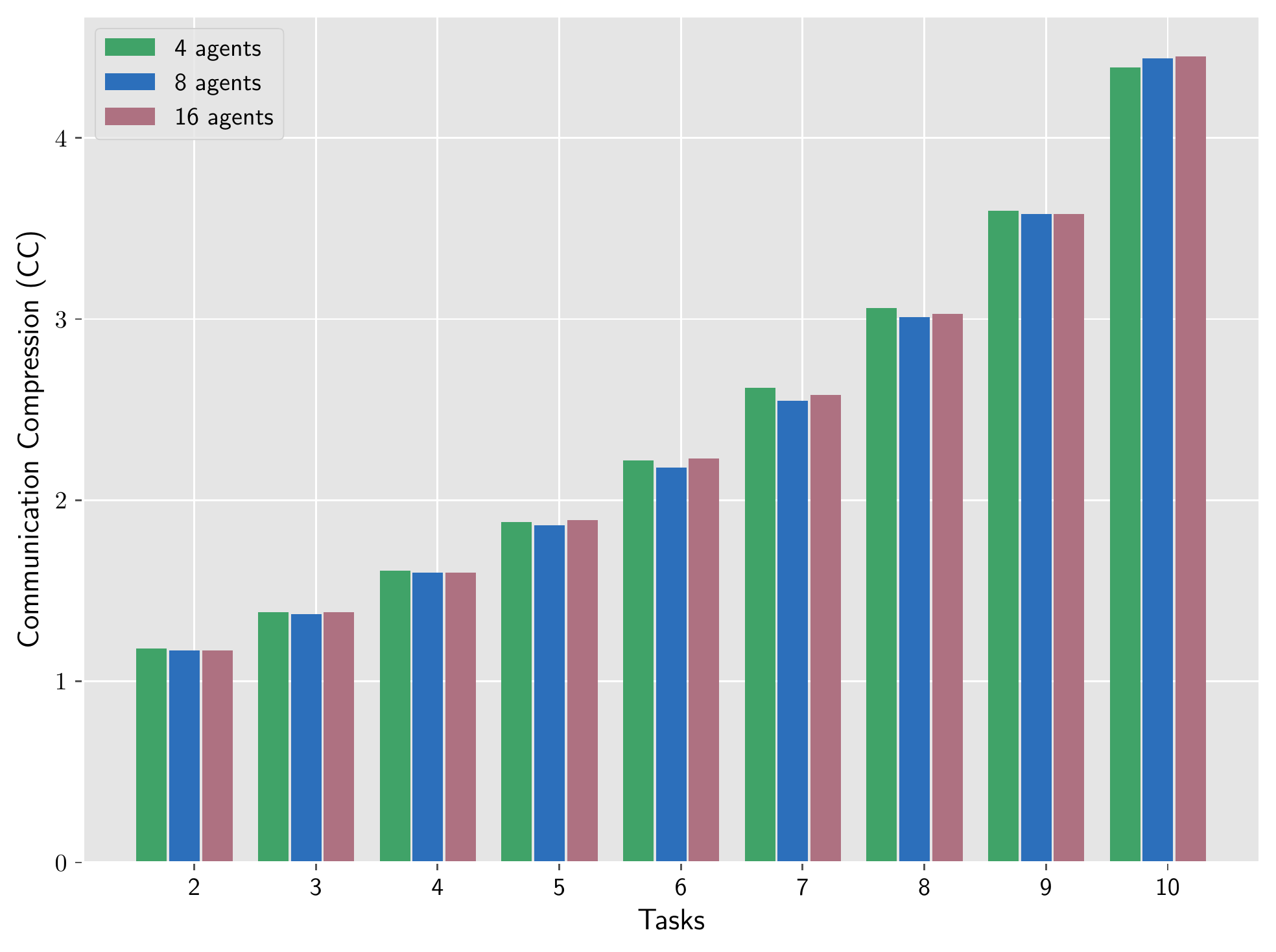}
  \vspace{-3mm}
  \caption{Task-wise CC for Split CIFAR-100 over AlexNet with ring topology}
  \label{fig:taskCCcifar}
\end{figure}

\begin{figure}[h!]
\centering
  \includegraphics[width=0.65\textwidth]{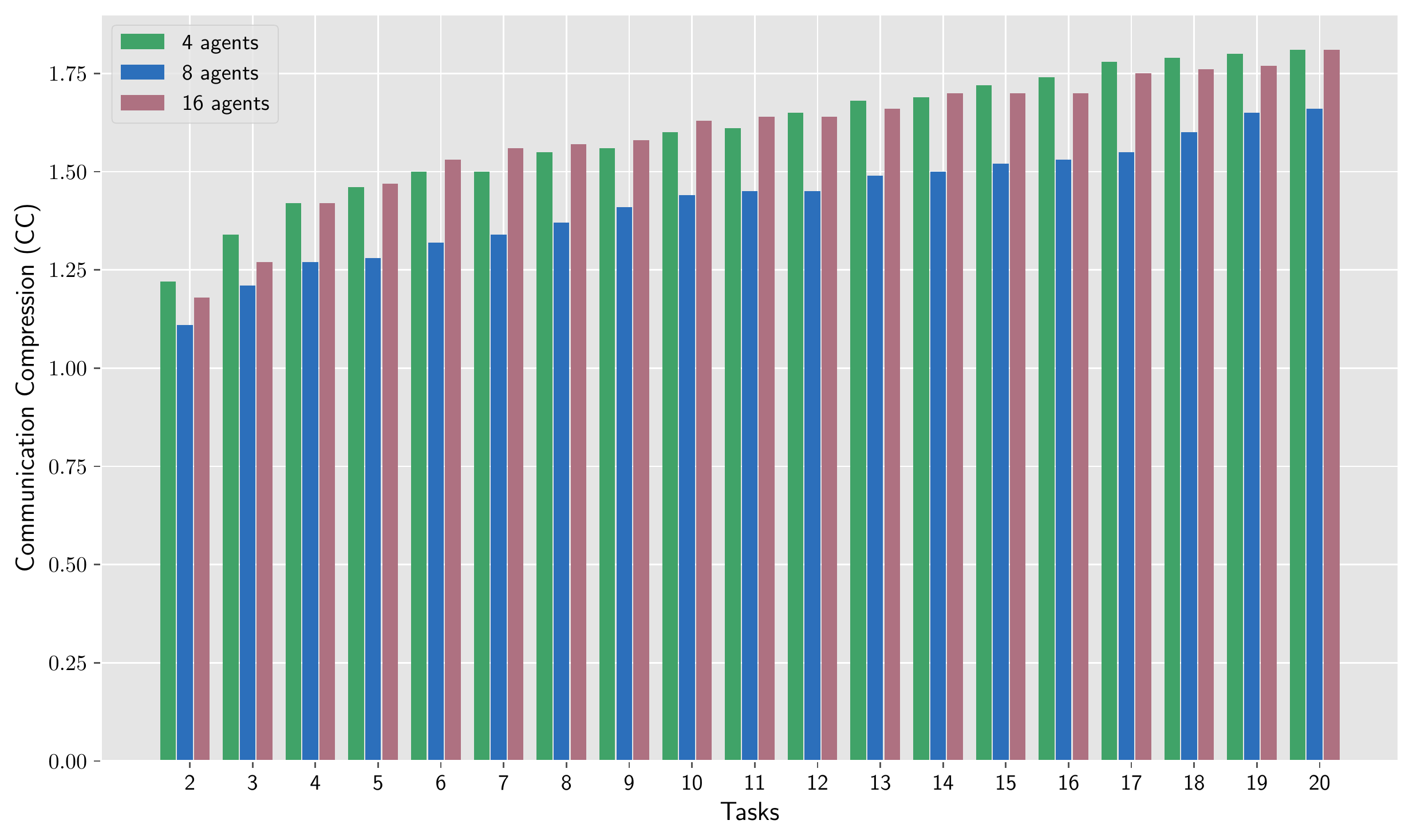}
  \vspace{-3mm}
  \caption{Task-wise CC for Split miniImageNet over ResNet-18 with ring topology}
  \label{fig:taskCCminii}
\end{figure}

\begin{figure}[!ht]
\centering
    \subfloat[\centering Task 2]{{\includegraphics[width=0.45\textwidth]{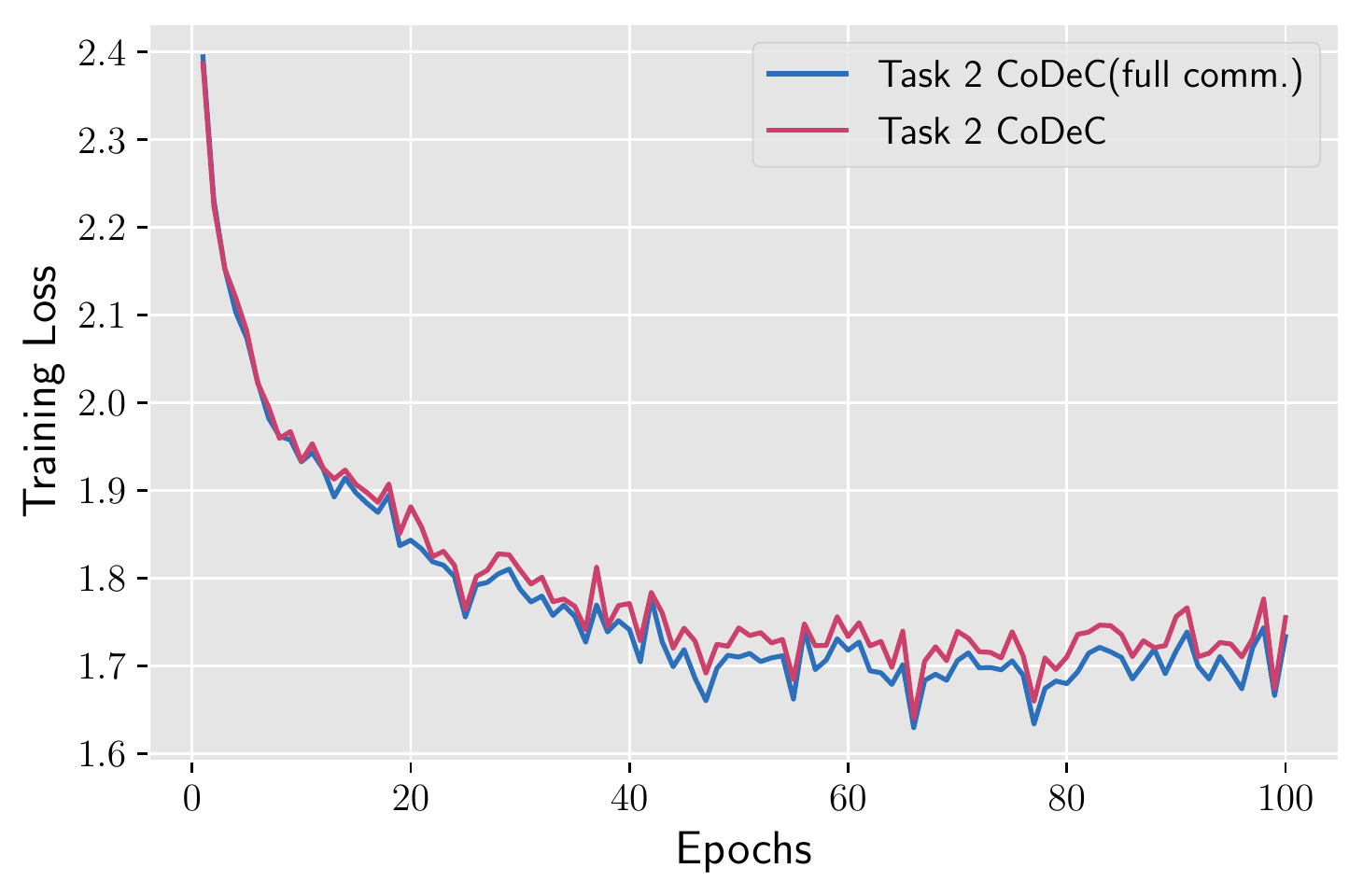} }}%
    \qquad
    \subfloat[\centering Task 9]{{\includegraphics[width=0.45\textwidth]{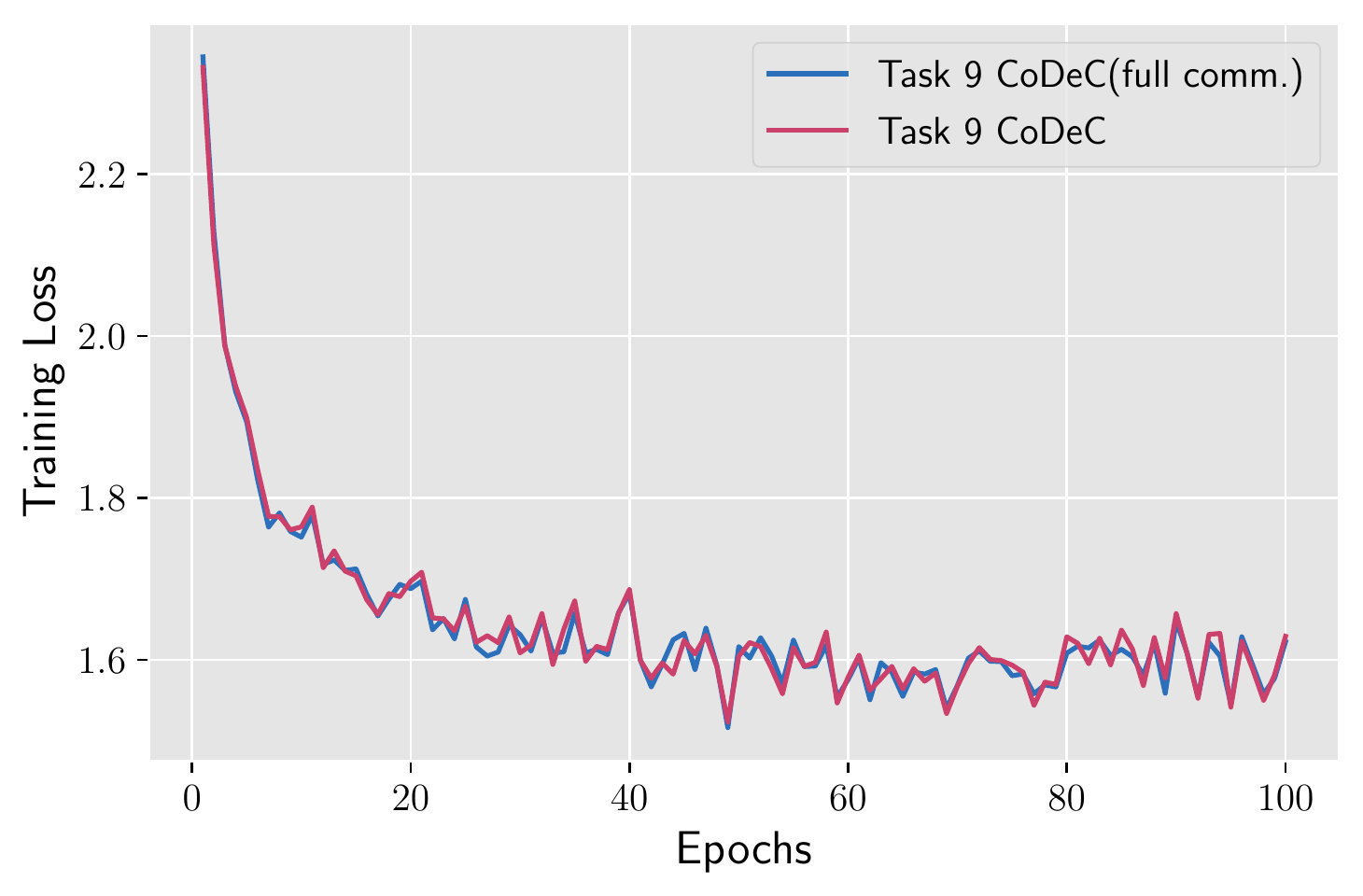} }}%
  \caption{Training loss vs epochs for (a) task 2 and (b) task 9 in Split CIFAR-100 sequence with CoDeC(full comm.) and CoDeC using AlexNet over a directed ring with 8 agents}
  \vspace{125in}
  \label{fig:trainloss}
\end{figure}

\noindent \textbf{Task-wise CC for Split CIFAR-100 and Split miniImageNet:} We present additional results for task-wise CC, similar to figure \ref{fig:taskCC}. Figure \ref{fig:taskCCcifar} shows task-wise CC ranging from 1.2x to 4.45x for Split CIFAR-100. Figure \ref{fig:taskCCminii} demonstrates that task-wise CC ranges from 1.2x to 1.8x for Split miniImageNet.
\vspace{4mm}
\newline \noindent \textbf{Training Loss vs Epochs with and without compression:} We present some results to emphasize the lossless nature of our proposed communication compression scheme. Figure \ref{fig:trainloss} shows training loss after each epoch for a particular agent for task 2 and 9 in Split CIFAR-100 sequence with and without compression. The convergence rate of the training loss is not affected by applying the proposed compression scheme.

\end{document}